\documentclass[lettersize,journal]{IEEEtran}
\ifCLASSOPTIONcompsoc
\usepackage[nocompress]{cite}
\else
\usepackage{cite}
\fi

\ifCLASSINFOpdf
\usepackage[pdftex]{graphicx}
\else
\usepackage[dvips]{graphicx}
\DeclareGraphicsExtensions{.eps}
\fi
\usepackage{amsmath,amsfonts}
\usepackage{array}
\usepackage{subfiles}
\usepackage{textcomp}
\usepackage{stfloats}
\usepackage{url}
\usepackage{verbatim}
\usepackage{graphicx}
\usepackage{cite}
\usepackage{tikz}
\usepackage{comment}
\usepackage{amsmath}
\usepackage{amssymb}
\usepackage{mathtools}
\usepackage{amsthm}
\usepackage{xspace}
\usepackage{bm}
\usepackage{amssymb}
\usepackage{pifont}
\usepackage[utf8]{inputenc} 
\usepackage[T1]{fontenc}    
\usepackage{url}            
\usepackage{booktabs}       
\usepackage{amsfonts}       
\usepackage{nicefrac}       
\usepackage{microtype}      
\usepackage{xcolor}         
\newcommand{\cmark}{\ding{51}\xspace}%
\newcommand{\xmark}{\ding{55}\xspace}%
\newcommand{\xmarkg}{\textcolor{lightgray}{\ding{55}}\xspace}%
\usepackage{multirow}
\usepackage[textsize=tiny]{todonotes}
\usepackage{colortbl}
\definecolor{mygray}{gray}{.9}
\definecolor{mypink}{rgb}{.99,.91,.95}
\definecolor{mycyan}{cmyk}{.1,0,0,0}

\usepackage{wrapfig}
\usepackage{caption}
\usepackage[linesnumbered,ruled]{algorithm2e}

\theoremstyle{definition}

\let\oldnl\nl
\newcommand{\nonl}{\renewcommand{\nl}{\let\nl\oldnl}}

\renewcommand\thesection{\Alph{section}}

\usepackage{multirow}
\usepackage[font=scriptsize]{subfig}

\usepackage{soul}
\usepackage{color, xcolor}
\usepackage{verbatim}

\soulregister\cite7 
\soulregister\eqref7
\usepackage{booktabs}

\newcommand{\innerproduct}[2]{\langle #1, #2 \rangle}

\definecolor{mygray}{gray}{.9}
\definecolor{mypink}{rgb}{.99,.91,.95}
\definecolor{mycyan}{cmyk}{.1,0,0,0}
\usepackage{verbatim}

\usepackage{algorithmic}
\usepackage{algorithm}

\newif\ifcomments
\commentstrue 
\usepackage{color}

\begin{document}

\title{Fine-grained Visual-Text Prompt-Driven Self-Training for Open-Vocabulary Object Detection
}
\author{Yanxin~Long, Jianhua~Han, Runhui~Huang, Hang~Xu, Yi~Zhu, Chunjing Xu, Xiaodan~Liang\IEEEauthorrefmark{1}
\IEEEcompsocitemizethanks{
        \emph{\IEEEauthorrefmark{1}Xiaodan Liang is the corresponding author.}
        
		\IEEEcompsocthanksitem Yanxin Long and Runhui Huang are with the School of Intelligent Systems Engineering, Shenzhen Campus of Sun Yat-sen University, Shenzhen 518107, China (e-mail: longyx9@mail2.sysu.edu.cn;
		huangrh9@mail2.sysu.edu.cn).
            \IEEEcompsocthanksitem Xiaodan Liang is with the School of Intelligent Systems Engineering, Shenzhen Campus of Sun Yat-sen University, Shenzhen 518107, China, and DarkMatter AI Research, Guangzhou 511458, China (e-mail: liangxd9@mail.sysu.edu.cn).
		
		
		Jianhua Han, Hang Xu, Yi Zhu and Chunjing Xu is with Huawei Noah'ark Lab, Shanghai 201206, China (e-mail: hanjianhua4@huawei.com);
		chromexbjxh@gmail.com; zhuyi215@mails.ucas.ac.cn; xuchunjing@huawei.com).}}

\maketitle
\renewcommand\thesection{\Roman{section}}


\begin{abstract}
Inspired by the success of vision-language methods~(VLMs) in zero-shot classification, recent works attempt to extend this line of work into object detection by leveraging the localization ability of pre-trained VLMs and generating pseudo labels for unseen classes in a self-training manner.
However, since the current VLMs are usually pre-trained with aligning sentence embedding with global image embedding, the direct use of them lacks fine-grained alignment for object instances, which is the core of detection.
In this paper, we propose a simple but effective fine-grained \textbf{V}isual-\textbf{T}ext \textbf{P}rompt-driven self-training paradigm for \textbf{O}pen-\textbf{V}ocabulary \textbf{D}etection~(\textbf{VTP-OVD}) that introduces a fine-grained visual-text prompt adapting stage to enhance the current self-training paradigm with a more powerful fine-grained alignment.
During the adapting stage, 
we enable VLM to obtain fine-grained alignment by using learnable text prompts to resolve an auxiliary dense pixel-wise prediction task.
Furthermore, we propose a
visual prompt module to provide the prior task information~(i.e., the categories need to be predicted) for the vision branch to better adapt the pre-trained VLM to the downstream 
tasks.
Experiments show that our method achieves the state-of-the-art performance for open-vocabulary object detection, e.g., 31.5\% mAP on unseen classes of COCO.
\end{abstract}

\section{Introduction}\label{sec:introduction}
\IEEEPARstart{T}{he} dominant object detection paradigm uses supervised learning to predict limited categories of the object. 
However, the existing object detection datasets usually contain only a few categories due to the time-consuming labeling procedure, e.g., 20 in PASCAL VOC~\cite{everingham2010pascal}, 80 in COCO~\cite{lin2014microsoft}. 
Some methods~\cite{gupta2019lvis} attempt to expand the categories to a larger scale by adding more labeled datasets. However, it is time-consuming and difficult to guarantee sufficient instances in each class due to the naturally long-tailed distribution.
On the other hand, previous methods~\cite{bansal2018zero,demirel2018zero,rahman2019transductive,hayat2020synthesizing} follow the setting of zero-shot detection and align the visual embeddings with the text embeddings generated from a pre-trained text encoder on base categories, but they still have a significant performance gap compared to the supervised counterpart.

Benefiting from the massive scale of datasets collected from the web~\cite{jia2021scaling, chen2020uniter}, recent vision-language pre-training models (VLM), e.g., CLIP~\cite{radford2021learning}, have shown a surprising zero-shot classification capability by aligning the image embeddings with corresponding caption.
However, it is a challenging research direction to transfer this zero-shot classification ability to object detection in dense prediction framework, because fine-grained pixel-level alignment, which is essential for dense tasks, is missing in the current visual language pre-training models.
\begin{figure}[t!]
\begin{center}
\includegraphics[ width=1.0\linewidth]{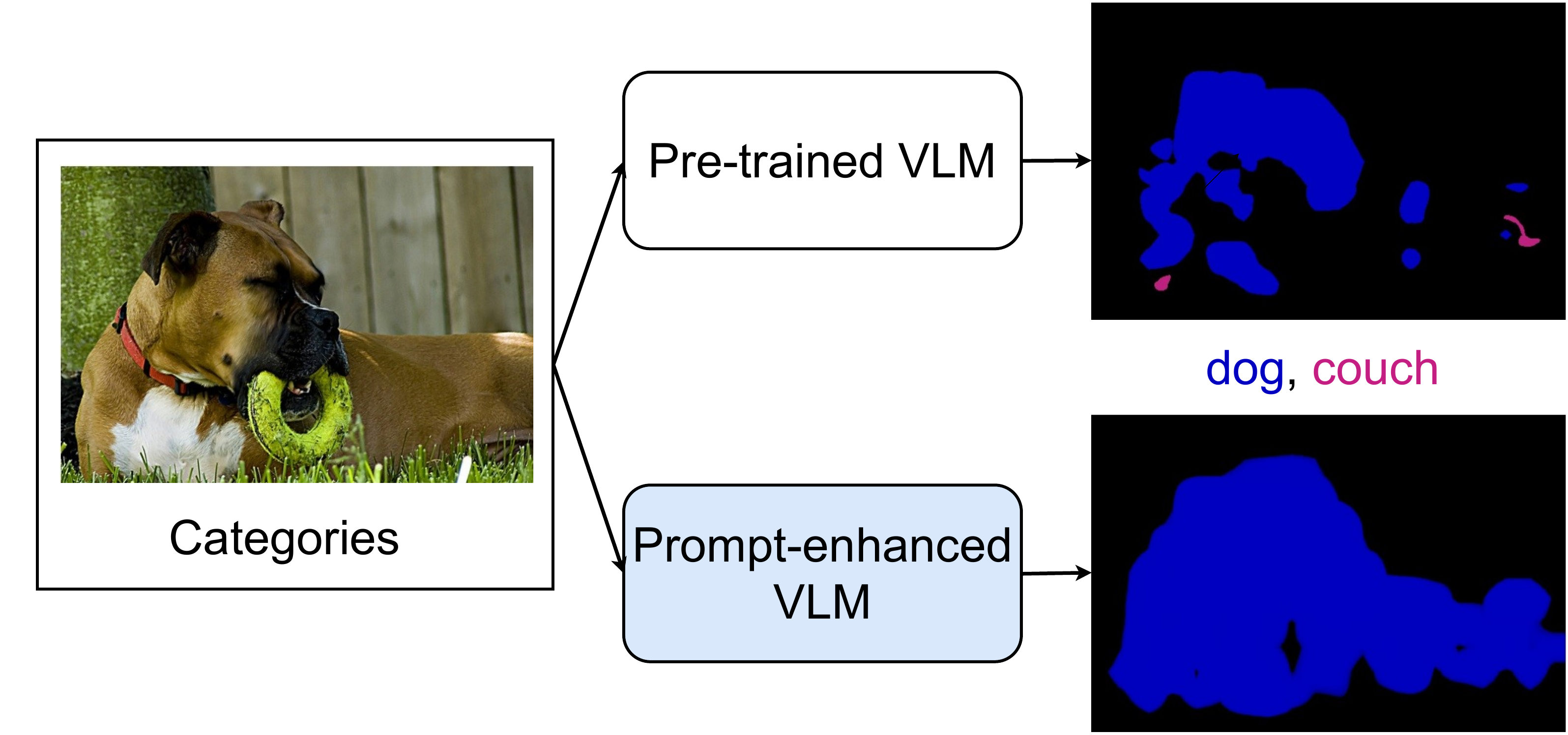}
\captionsetup{width=1.0\linewidth}
    \caption{Illustration of the dense score map of our proposed prompt-enhanced VLM compared with the original VLM.   }
    \vspace{-6mm}
    \label{fig:dense_align}
    \end{center}
\end{figure}


There exist some attempts to train an open-vocabulary detector by leveraging the zero-shot classification ability of a pre-trained VLM. 
The work~\cite{gao2021towards} proposes a basic self-training pipeline, which utilizes the activation map of the noun tokens in the caption from the pre-trained VLM and generates the pseudo bounding-box label.
However, existing OVD methods just take advantage of the global text-image alignment ability of VLMs, thus failing to capture dense text-pixel alignment and severely hindering the self-training performance of OVD tasks. 
Directly using them by activation map cannot fully adapt to downstream detection tasks, which require better dense representations.
For example, as shown in the upper part of Fig.~\ref{fig:dense_align}, directly utilizing the pre-trained VLM can only obtain a low-quality and incomplete dense score map for the dog in the input image, which is harmful for the next pseudo labeling stage. 
Recent works~\cite{zhou2022detecting,feng2022promptdet} attempt to build generic object detection frameworks by scaling to larger label spaces, while they are costly required to acquire large-scale annotations from bigger datasets.

\begin{figure}[t!]
		\begin{center}

\includegraphics[width=0.98\linewidth]{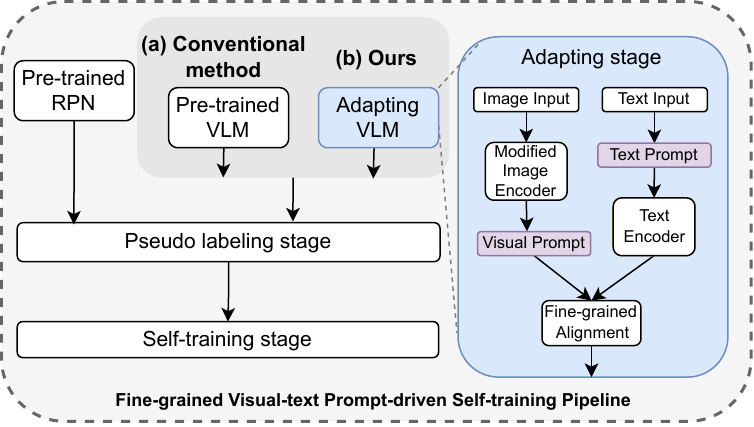}
		\end{center}
		\caption{
  Illustration of our proposed prompt-driven self-training pipeline VTP-OVD compared with the conventional self-training pipeline for open-vocabulary object detection. We introduce an additional adapting stage of VLM to obtain fine-grained alignment via adopting learnable visual and text prompt modules in image encoder and text encoder. }
		\label{fig:intro_figure}
\end{figure}

To achieve fine-grained alignment over dense pixels and avoid extra data cost, we propose \textbf{VTP-OVD}, a \textbf{V}isual-\textbf{T}ext \textbf{P}rompt-driven self-training paradigm for \textbf{O}pen-\textbf{V}ocabulary \textbf{D}etection~(\textbf{VTP-OVD}), to improve the robustness and generalization capability. 
Inspired by the recent advance in learning-based prompting methods~\cite{liu2021pre,zhou2021learning} from Natural Language Processing (NLP) community,  we design a novel prompt-driven self-training framework (Fig.\ref{fig:intro_figure}) for better adaption of the pre-trained VLM to new detection tasks.
In detail, VTP-OVD designs a fine-grained visual-text prompt adapting stage to obtain more powerful dense alignments for better pseudo label generation by introducing an additional dense prediction task.
Specifically in adapting stage, to reduce the domain gap of upstream and downstream tasks and obtain the semantic-aware visual embedding, we introduce the visual and text prompt modules into the learnable image encoder and text encoder of the original VLM respectively.
The text prompts provide dense alignment task cues to enhance category embedding, and the visual prompt module aligns enhanced categories information to each pixel.

Furthermore, with the prompt-enhanced VLM after adapting stage, a better pseudo label generation strategy is proposed for novel classes by leveraging the non-base categories' names as the label input and aligning each connect region of score map for each category.
As far as we know, our VTP-OVD is the first work to employ a learnable prompt-driven adapting stage in OVD task to capture fine-grained pixel-wise alignment and therefore to generate better pseudo labels.

We evaluate our VTP-OVD on the popular object detection dataset COCO~\cite{lin2014microsoft} under the well-known zero-shot setting, and further validate its generalization capability on PASCAL VOC~\cite{everingham2010pascal}, Objects365~\cite{shao2019objects365} and LVIS~\cite{gupta2019lvis} benchmarks.
Experiment results show that our VTP-OVD achieves the state-of-the-art performance on detecting novel classes~(without annotations), e.g., 31.5\% mAP on the COCO dataset.
Besides, VTP-OVD also outperforms other open-vocabulary detection~(OVD) methods when directly adapting the model trained on COCO to perform open-vocabulary detection on the three other object detection datasets.
To demonstrate the effectiveness of the two learnable prompt modules, further experiments are conducted to show that fine-tuned with these two modules, VTP-OVD can generate a better dense score map (4.3\% mIoU higher) compared to directing using the pre-trained CLIP~\cite{radford2021learning} model. 

\vspace{-2mm}
\section{Related Work}

\begin{figure*}[t!]
		\begin{center}
            \includegraphics[ width=0.9\linewidth]{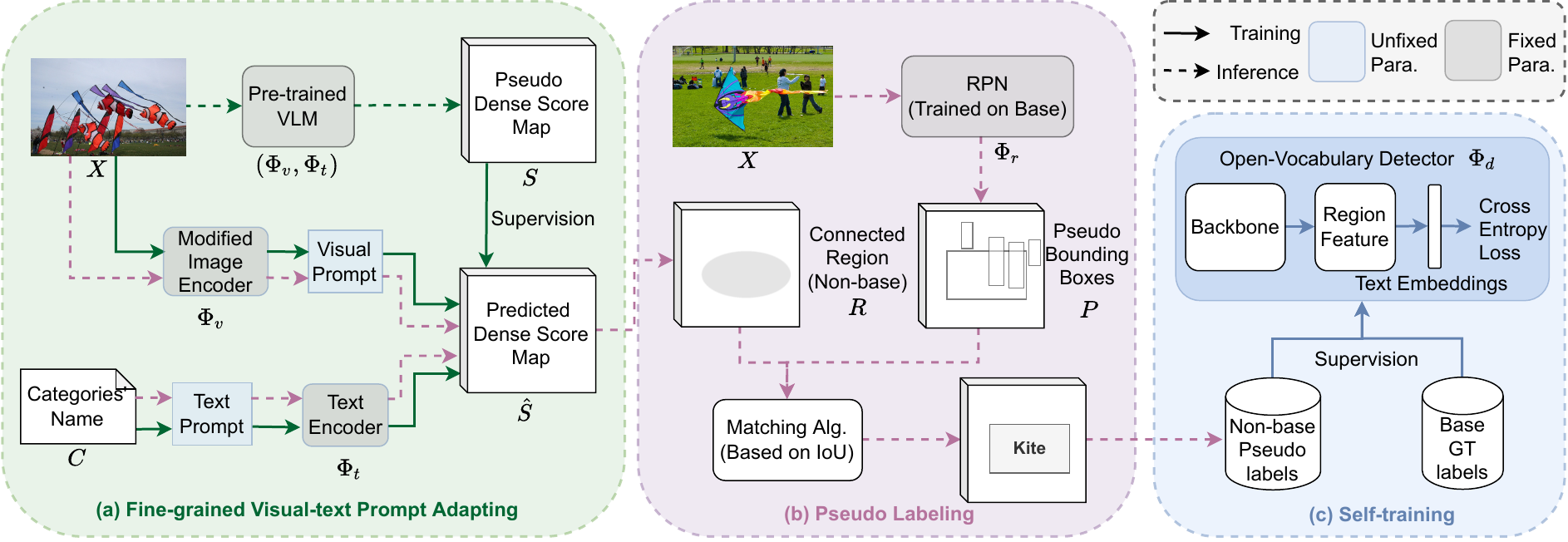}
		\end{center}
		\caption{Overview of proposed open-vocabulary detection pipeline VTP-OVD. VTP-OVD consists of three stages: (a) \textbf{Fine-grained visual-text prompt adapting stage} to improve the dense pixel-wise alignment via learnable vision-text prompts to obtain better pseudo bounding-box labels. (b) \textbf{Pseudo labeling stage} to obtain non-base classes' pseudo labels by utilizing the dense classification ability of prompt-enhanced VLM and the localization ability from a pre-trained RPN. (c) \textbf{Self-training stage} to feed the pseudo labels into an open-vocabulary detector for further training.  
		Different line color denotes different stages.
		}
		\label{fig:framework}
\end{figure*}

\subsection{Zero-shot Object Detection} Most zero-shot detection~(ZSD) methods~\cite{bansal2018zero,demirel2018zero,rahman2019transductive,hayat2020synthesizing} align the visual embeddings to the text embeddings of the corresponding base category generated from a pre-trained text encoder.
Several methods introduce GCN~\cite{9153181}, contrastive learning~\cite{yan2022semantics}, and a novel `polarity loss'~\cite{9812473} to bridge the gap between the visual embeddings and text embeddings.
Inspired by the success of VLMs~\cite{radford2021learning,li2021align}, several methods attempt to perform ZSD by training with image captions or directly leveraging a pre-trained VLM. 
\cite{zareian2021open} propose to pre-train the CNN backbone and a Vision to Language Module~(V2L) via grounding task on an image caption dataset, and then the whole architecture is fine-tuned with an additional RPN module. 
However, it still suffers from a large performance gap with the SOTA, and the domain of upstream and downstream datasets must remain similar to maintain performance~\cite{gao2021towards}. Based on a pre-trained VLM, \cite{gu2021open} distills the learned image embeddings of the cropped proposal regions to a student detector, \cite{bangalath2022bridging} further introduces an inter-embedding relationship matching loss to instill inter-embedding relationships. While each proposal needs to be fed forward into the image encoder of VLM. which requires huge computation costs. 
~\cite{gao2021towards} proposes a basic self-training pipeline based on ALBEF~\cite{li2021align}, which firstly utilizes Grad-CAM~\cite{selvaraju2017grad} to obtain dense activation region of specific words in the caption and then generate the pseudo bounding-box label by selecting the proposal that has the largest overlap with the activation region.
However, 
most VLMs lack the ability to perform pixel-wise classification since it is pre-trained with the correspondence of text embedding and global visual token instead of pixel embeddings.
Besides, the reliance on caption data limits the application of this generalization to some datasets without caption. Inspired by~\cite{9709514}, our method can obtain the pseudo annotations with the pixel-alignment patches.

\subsection{Vision-language Pre-training} 
The pre-training tasks of Vision-and-Language Pre-training (VLP) models can be divided into two categories: image-text contrastive learning
tasks and Language Modeling (LM) based tasks. The first category, e.g., 
CLIP \cite{radford2021learning} and ALIGN \cite{jia2021scaling}, aims to align the visual feature with textual feature in a cross-modal common semantic space.
The other category, e.g.,
VisualBERT \cite{li2019visualbert},
UNITER \cite{chen2020uniter}, M6 \cite{lin2021m6}, DALL-E \cite{ramesh2021zeroshot}, ERNIE-ViLG \cite{zhang2021ernie}, employs LM-like objects, include both autoregressive LM (e.g., image captioning~\cite{8331856}, VQA~\cite{8334194}) and masked LM (e.g., Masked Language/Region Modeling). 
Different from previous works that only focus on global feature alignment, our proposed model can achieve pixel-aware alignments via prompts and
perform better in downstream tasks of pixel-wise dense prediction.

\subsection{Prompt Tuning}
Freezing the pre-trained models with only tuning the soft prompts can benefit from efficient serving and matching the performance of full model tuning.
Prompt tuning has been verified as an effective method to mitigate the gap between pre-training and fine-tuning. As a rapidly emerging field in NLP
\cite{liu2021pre}, prompt tuning is originally designed for probing knowledge in pre-trained language models \cite{petroni2019language} and now applied in various NLP tasks, e.g., language understanding \cite{schick2020s}, emotion detection \cite{9881877} and generation \cite{li2021prefix}. 
Prompt tuning has now been extended to vision-language models. Instead of constructing hand-crafted prompts in CLIP \cite{radford2021learning}, CoOp \cite{zhou2021learning} proposes tuning soft prompts with unified context and class-specific context in downstream classification task. CPT \cite{yao2021cpt} proposes colorful cross-modal prompt tuning to explicitly align natural language to fine-grained image regions. 
The main differences between our usage of learnable multi-modal prompts with the previous prompt tuning works lie in three aspects: 1) Unaligned upstream and downstream tasks; 2) Multi-modal; 3) Learnable prompts for self-training-based open-vocabulary object detection.

\vspace{-2mm}
\section{Method}

In this section, we first briefly introduce the VTP-OVD framework.
Then we describe the details of different stages in VTP-OVD: 
a) learnable multi-modal prompts in fine-grained alignment stage,  which are used to enable VL model to obtain fine-grained pixel-wise alignment ability;
b) better pseudo label generation strategy in the pseudo labeling stage and c) the details of the final self-training stage.

\noindent\textbf{Basic Notations}: 
We construct a combined categories set (denoted as $C$) via extracting the categories from several large-scale detection datasets. 
The base~(seen) categories, novel~(unseen) categories, and non-base categories are denoted as $C_B$, $C_N$, $C_{\overline{B}}$ (\textit{i.e.}, $C$ = $\{C_B, C_{\overline{B}}\}, C_N \in C_{\overline{B}}$).
The input image is represented as $X$.
We use $\Phi_v$ and $\Phi_t$ to represent the modified image encoder~\cite{zhou2021denseclip} and text encoder of the original VLM specifically.

\vspace{-4mm}
\subsection{VTP-OVD Framework}
\label{sec:framework}


Fig.~\ref{fig:framework} illustrates the overall pipeline of the proposed VTP-OVD. We use different colors to indicate the parameter fixing or requiring training as clarified in the upper right corner of Fig.~\ref{fig:framework}.
Firstly, at the 
fine-grained visual-text prompt adapting stage, by employing the learnable text prompt module $prompt_t$ and visual prompt module $prompt_v$ into the text encoder $\Phi_t$ and image encoder $\Phi_v$, we can fine-tune the whole network under dense supervision of pseudo dense score map with other parameters fixed. 
Secondly, for pseudo labeling stage, we obtain the pseudo labels of non-base
classes by leveraging the dense classification ability of the fine-tuned VL model and the location ability of a pre-trained (on base classes) Region Proposal Network~(RPN) $\Phi_r$.
Finally, in the final self-training stage, an open-vocabulary detector $\Phi_d$ is further trained with the combination of the ground truth of base classes and the pseudo labels of 
non-base
classes to fulfill the self-training pipeline. The complete training procedure can be found in Algorithm \ref{alg:pseudo code}.
\subsection{Adapting Stage Via Multi-Modal Prompts}
\label{sec:finetuning}

The crucial part of the self-training pipeline for open-vocabulary detection~(OVD) lies in the pseudo labeling stage that determines the final detection performance of novel classes.
To generate better pseudo labels for the detection task, 
our method first aligns each pixel with a category and then adopts a pre-trained RPN for bounding-box localization. 
Since most vision-language methods~(VLM) are trained via the alignment between the whole image and the corresponding caption, they lack the dense alignment ability between pixels and categories. 
Thus in this section, we focus on modifying the pre-trained vision-language model~(VLM) to enhance the current self-training paradigms with fine-grained adapting stage via a newly designed dense alignment loss function and learnable text/visual prompts.
Note that to obtain the dense pixel-level visual embeddings instead of the global visual embedding, following \cite{zhou2021denseclip}, we modified the image encoder by removing the query and key embedding layers, which is denoted as $\Phi_v$ in Fig.~\ref{fig:framework} (a).

\begin{figure}[t!]
		\begin{center}
\includegraphics[width=1\linewidth]{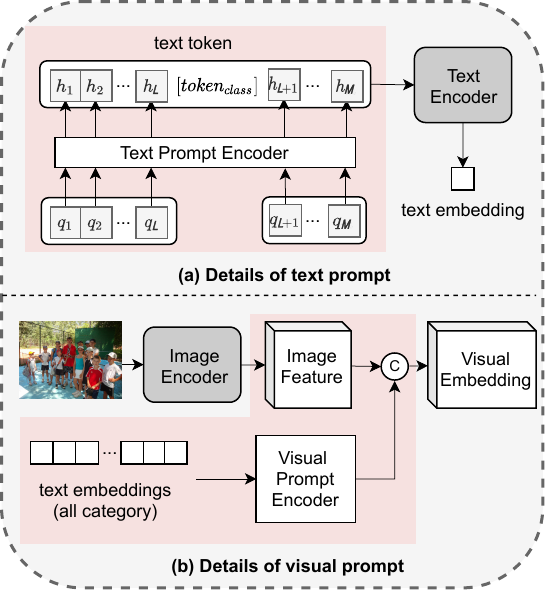}

		\end{center}
		\caption{
        Details of the visual prompt and text prompt. The text prompt encoder consists of a long-short term memory network (LSTM) and two-layer multilayer perceptron (MLP) to adapt to the dense task. The visual prompt encoder consists of a cross-attention module to obtain the semantic-aware visual embeddings. `c' represents the concatenation operation. 
		}
		\label{fig:details_prompt}
\end{figure}

\vspace{3mm}

\subsubsection{Dense Alignment Task}

As shown in Fig. \ref{fig:framework}, we introduce a dense alignment task to fine-tune the VLM under the supervision of pseudo dense score map $S\in \mathbb{R}^{HW \times |C|}$, where $C$ denotes the all categories, including multi dataset categories, 
and $|C|$ represents the number of $C$.
The $S$ is generated by calculating the similarity of pixel-wise visual embeddings of input image $X\in \mathbb{R}^{HW}$ with each category via the original VLM.

The dense score map $S$ is calculated by:

\begin{equation}
\label{eq:seg map}
V = \Phi_v(X), T = \Phi_t(C), S = \innerproduct{\frac{V}{\Vert V \Vert}}{\frac{T}{\Vert T \Vert}}
\end{equation}

where $\innerproduct{\cdot}{\cdot}$ denotes the inner product operation, $\Vert\cdot\Vert$ denotes the normalization operation. $V \in  \mathbb{R}^{HW \times D}$ and $T \in \mathbb{R}^{|C| \times D}$ represent the visual embeddings and text embeddings specifically.
To reduce the domain gap of upstream and downstream tasks and obtain the semantic-aware visual embedding, we further introduce the visual and text prompt modules into the learnable image encoder and text encoder of the original VLM respectively. 
Benefit from the same task form as upstream tasks, most previous prompt-based tuning works~\cite{liu2021pre, gao2021clip,zhou2021learning} still performs well on downstream tasks with different data domains. However, our VTP-OVD utilizes learnable multi-modal prompts to obtain the fine-grained feature alignment, which is different from the upstream global alignment in the task domain.
The objective of 
dense alignment stage is:

\begin{equation}
     \mathcal{L}_{FT} = {L}_{CE}(S, \hat{S}),
\end{equation}
where ${L}_{CE}$ represents the cross-entropy loss.
Similar to Eq. (\ref{eq:seg map}), the model prediction $\hat{S}\in \mathbb{R}^{HW \times C}$ with the image input $X$ is calculated as:

\begin{equation}
\label{eq:ft seg map}
\hat{S} = \innerproduct{\frac{V'}{\Vert V' \Vert}}{\frac{T'}{\Vert T' \Vert}}
\end{equation}
where $V' \in  \mathbb{R}^{HW \times D}$ and $T' \in \mathbb{R}^{|C| \times D}$  denote the image feature and text feature integreated with prompt embeddings.

\subsubsection{Text Prompt}
\label{sec:text prompt}

Prompt tuning has been verified as an effective method to probe knowledge in pre-trained language models \cite{petroni2019language} and applied in various NLP tasks \cite{schick2020s,li2021prefix}. For example, the hand-crafted text prompts, e.g., `a photo of a \{\}.', have been adopted to adapt the pre-trained VLM to the different downstream tasks
~\cite{radford2021learning,gu2021open}. Inspired by these works, we further introduce a learnable text prompt encoder to provide dense alignment task cues for enhancing categories embeddings since downstream text-pixel alignment task is quite different from the upstream global text-image alignment task.

As shown in Fig.~\ref{fig:details_prompt} (a), the text prompt encoder maps the learnable prompt tokens $Q \in \mathbb{R}^{M \times D}$ into the prompt embeddings $ H \in \mathbb{R}^{M \times D}$. $M$ denotes the total number of prompt tokens.
We believe that the $h_i \in H$ should depend on each other since the words in a natural sentence always have a strong contextual relationship.
Therefore for the structure of the proposed text prompt encoder module, it contains a bidirectional long-short term memory network~(LSTM) to get the contextual information followed by a ReLU activated two-layer multilayer perceptron (MLP)~\cite{liu2021gpt}:

\begin{equation}
    H  = MLP(LSTM(Q))
\end{equation}


After that, we concatenate the same text prompt embeddings with each class token $c \in C$ to obtain the prompt-enhanced text embedding $t' \in T'$ by:
\begin{equation}
    t' = \Phi_t ( [\{h_i\}_{i=0}^L, c,  \{h_i\}_{i=L+1}^M]),
\end{equation}
where $[\cdot]$ denotes the concatenation operation, and $L$ denotes the length of prompt tokens added before the category token.

\subsubsection{Visual Prompt}
\label{sec:visual prompt}
As shown in Fig.~\ref{fig:details_prompt}~(b), the text prompts employ dense alignment task cues to enhance category embedding. Therefore, to align the enhanced categories information to each pixel, a visual prompt encoder is built upon the cross-attention mechanism to generate semantic-aware visual embedding for each pixel to improve dense alignment. Similar to ActionCLIP~\cite{wang2021actionclip}, which utilizes positional embeddings as visual prompt to provide additional temporal information, ours is also formed as a post-network prompt module~\cite{wang2021actionclip} to provide additional semantic cues and we only finetune the visual prompt module while fixing the visual encoder.
By taking each pixel embedding in visual features $V = \Phi_v(X)\in \mathbb{R}^{HW \times D}$ 
as the input of the query and the text embeddings of all categories $T' \in \mathbb{R}^{|C| \times D}$ as the input of key and value, the cross-attention block output the semantic-aware visual prompts $\Tilde{V} \in \mathbb{R}^{HW \times D}$ for each pixel. Then they're concatenated with the visual features followed by an 1-layer multilayer perceptron (MLP) to obtain the semantic-aware visual embeddings $V' \in \mathbb{R}^{HW \times D}$ ~(see Fig.~\ref{fig:details_prompt} (b)):


\begin{equation}
\begin{aligned}
    V' & = MLP(V \oplus \Tilde{V}) \\
    \Tilde{V} &= softmax(\frac{V \times T'^{\mathsf{T}}}{\sqrt{D}}) \cdot T'
\end{aligned}
\end{equation}

where $\oplus$ and $D$ represent the concatenation operation and the feature dimension of text embeddings $T'$.
$MLP$ denotes the 1-layer multi-layer perceptron 
to reduce back the feature dimension.

\begin{algorithm}[t!]
   \caption{VTP-OVD pseudo code}
   \label{alg:pseudo code}
\begin{algorithmic}
   \STATE {\bfseries Input:} Images $X$, Base categories $C_B$, Non-base classes $C_{\overline{B}}$, Image encoder $\Phi_v$, Text encoder $\Phi_t$, Pretrained RPN model $\Phi_{r}$, Open-vocabulary detector $\Phi_d$, Visual prompt $prompt_v$, Text prompt $prompt_t$, Fine-tuning iter $Iter_{f}$, Training iter $Iter_{t}$, Inner product function $\innerproduct{\cdot}{\cdot}$.
   
   \STATE $\Phi_v, \Phi_t \leftarrow$ initialized w/ CLIP pre-trained
   
   \STATE \# \textbf{1. Fine-grained visual-text prompt adapt stage}
   \STATE $C \leftarrow$ all classes $C_B + C_{\overline{B}}$
   \STATE $V, T \leftarrow$ image feature and text feature $\Phi_v(X), \Phi_t(C)$
   \STATE $S \leftarrow$ pseudo dense score map
   $\innerproduct{\frac{V}{\Vert V \Vert}}{\frac{T}{\Vert T \Vert}}$
   \STATE $V' \leftarrow$ add random-initialed $prompt_v$
   \STATE $T' \leftarrow$ add random-initialed $prompt_t$
   \FOR{$t=1$ {\bfseries to} $Iter_{f}$}
    \STATE $\hat{S} \leftarrow$ model prediction $\innerproduct{\frac{V'}{\Vert V' \Vert}}{\frac{T'}{\Vert T' \Vert}}$
    \STATE $\mathcal{L}_{FT} \leftarrow$ cross-entropy loss ${L}_{CE}(S, \hat{S})$
    \STATE $V', T' \leftarrow$ updated by $\mathcal{L}_{FT}$
   \ENDFOR
   
   \STATE \# \textbf{2. Pseudo labeling stage}
   \STATE $P \leftarrow$ generated proposals $\Phi_{r}(X)$
   \STATE $
   \hat{S}_{\overline{B}} \leftarrow$ pseudo non-base dense score map
   $\innerproduct{\frac{V'}{\Vert V' \Vert}}{\frac{T^{'}_{\overline{B}}}{\Vert T^{'}_{\overline{B}} \Vert}}$
   \STATE $R \leftarrow$ connected regions of non-base categories $\hat{S}_{\overline{B}}$
   \STATE pseudo labels $\leftarrow IoU(R, P) < \gamma $
   
   \STATE \# \textbf{3. Self-training stage}
   \FOR{$t=1$ {\bfseries to} $Iter_{t}$}
   \STATE $\hat{Y} \leftarrow$ model prediction $\Phi_d(X)$
   \STATE $Y \leftarrow$ ground truths~(for $C_B$) and pseudo labels~(for $C_{\overline{B}}$)
   \STATE $\mathcal{L}_T \leftarrow$  $\mathcal{L}_{CLS}(Y, \hat{Y})$ + $\mathcal{L}_{REG}(Y, \hat{Y})$
   \STATE $\Phi_d \leftarrow$ updated by $\mathcal{L}_T$
   \ENDFOR
   
\end{algorithmic}
\end{algorithm}

\subsection{Pseudo Label Generation}
\label{sec:pseudo labeling}
Based on the more precise dense score map generated by the fine-tuned VLM with learnable text and visual prompts, we can obtain the
non-base
classes' pseudo labels by additionally leveraging the location ability of a pre-trained region proposal network~(RPN) $\Phi_{r}$~(shown in Fig. \ref{fig:framework}(b)).
Note that $\Phi_{r}$ is only trained on base classes, and the experiments in~\cite{gu2021open} already demonstrate that training only on the base categories can achieve comparable recall to average recall of the overall categories.
Previous pseudo labeling generation strategy~\cite{gao2021towards} used the objects of interest in the caption, which not only harms the transfer ability to the detection datasets without the caption but also is limited to the uncompleted description of caption data.
For example, such a strategy cannot generate the pseudo labels of classes that are not included in the caption.

To address these issues, We
directly use 
non-base
classes' names as the input of fine-tuned text encoder $\Phi_t$.
Besides, to better distinguish the background classes, instead of directly using the word "background", we 
treat all the base classes as background since we already have the ground-truth annotation of these classes.
Then we obtain the pseudo dense score map of 
non-base
classes $\hat{S}_{\overline{B}}\in \mathbb{R}^{HW \times |C_{\overline{B}}|}$ from the fine-tuned VLM with the procedure same to Eq.~(\ref{eq:ft seg map}).  
For each image $x_i$, we firstly compute the connected regions $r_j\in \mathbb{R}^{HW}$ for $j$-$th$ 
non-base
category on $\hat{S}_{\overline{B}}$ by setting a similarity threshold $\delta$, then we adopt the intersection of union~(IoU) of $k$-$th$ proposal $p_i^k$ from RPN and $r_j$ as confidence score:

\begin{equation}
    score_{i, j, k} = IoU(p_i^k, r_j)
\end{equation}
where $p_i = \Phi_r(x_i)$ and the $score_{i, j, k}$ denotes the confidence score that the proposal $p_k$ belongs to the $j$-${th}$ 
non-base
category for image $x_i$.
Finally, a hard score threshold $\gamma$ is adopted on the confidence score to filter out pseudo boxes with less confidence.

\noindent\textbf{Self-training stage.}
After obtaining the pseudo bounding-box labels for 
non-base
classes, together with the ground-truth annotations of base classes, we are able to train a final open-vocabulary detector $\Phi_d$ to fulfill the self-training pipeline~(shown in Fig. \ref{fig:framework}(c)) via Faster-RCNN~\cite{ren2015faster}.
We build the $\Phi_d$ by replacing the last classification layer with the text embeddings generated by the text encoder on all categories $V'$.
The objective of self-training stage $\mathcal{L}_T$ is calculated as:

\begin{equation}
    \mathcal{L}_T = \mathcal{L}_{CLS}(Y, \hat{Y}) + \mathcal{L}_{REG}(Y, \hat{Y})
\end{equation}
Where the ${L}_{CLS}$ and ${L}_{REG}$ denote the cross-entropy loss for classification head and L1-loss for regression head, $\hat{Y}$ and $Y$ represent the model prediction $\Phi_d(X)$ and annotations combined with pseudo labels of 
non-base
classes and the ground-truth of base classes.
The trained detector $\Phi_d$ can then transfer to other detection datasets by providing the class names. 

\vspace{-4mm}
\section{Experiments}

\subsection{Open-Vocabulary Object Detection Setups}

\textbf{Benchmark Setting}:
We benchmark on the widely-used object detection dataset COCO 2017~\cite{lin2014microsoft}.
Following the well-known settings~\cite{bansal2018zero} adopted by many OVD methods, we divide the categories used in COCO into 48 base~(seen) classes and 17 novel~(unseen) classes. 
Each method can only be trained on annotations of the base categories and then predicts the novel categories without seeing any annotations of these categories.
We report the detection performance of both the base and novel classes during inference, as the generalized settings used in~\cite{zareian2021open}. 
We also evaluate the generalization ability of the trained detector by directly transferring it to three other object detection datasets, including PASCAL VOC~\cite{everingham2010pascal}, LVIS~\cite{gupta2019lvis} and Objects365~\cite{shao2019objects365}.  

\textbf{Evaluation Metric}:
We use mean average precision~(mAP) with IoU threshold 0.5 as the evaluation metric. We pay more attention to the novel class performance since we aim to build an open-vocabulary detector and the annotations of base classes are already provided.

\begin{table*}[t!]
\caption{Open-vocabulary detection performance comparison on COCO datasets. VTP-OVD (w/o FT) denotes the VTP-OVD without fine-grained adapting stage. We can observe that VTP-OVD achieves the state-of-the-art detection performance on the novel classes. Without the fine-tuning stage, the performance on the novel classes suffers a large drop. Note that OVR-CNN pre-trains a self-designed vision-language model with a vision-to-language~(V2L) module by itself. ViLD* trains the detector with data augmentations of large-scale jittering (LSJ)~\cite{ghiasi2021simple} and longer training schedule (16x). mAP~(\%) is reported. }
\label{table:main_results}
\vskip 0.15in
\begin{center}
\begin{small}
\begin{sc}
\begin{tabular}{l|cc|ccc}
\hline
\noalign{\smallskip}
Method & VL Model & Using COCO caption &  Novel & {\color{gray}Base} & {\color{gray}Overall} \\
\noalign{\smallskip}
\hline
\noalign{\smallskip}
SB\cite{bansal2018zero} & \xmark& \xmark & 0.31 & {\color{gray}29.2} & {\color{gray}24.9} \\
LAB\cite{bansal2018zero} & \xmark & \xmark & 0.22 & {\color{gray}20.8} & {\color{gray}18.0} \\
DSES\cite{bansal2018zero} & \xmark & \xmark & 0.27 & {\color{gray}26.7} & {\color{gray}22.1} \\
DELO\cite{zhu2020don} & \xmark & \xmark & 3.41 & {\color{gray}13.8} & {\color{gray}13.0} \\
PL\cite{rahman2020improved} & \xmark & \xmark & 4.12 & {\color{gray}35.9} & {\color{gray}27.9} \\
\noalign{\smallskip}
\hline
\noalign{\smallskip}
OVR-CNN \cite{zareian2021open}   & - & \cmark & 22.8& {\color{gray} 46.0}& {\color{gray}39.9}\\
ViLD* \cite{gu2021open} & CLIP& \xmark & 27.6 & {\color{gray}59.5} & {\color{gray}51.3} \\
\noalign{\smallskip}
\hline
\noalign{\smallskip}
\multicolumn{6}{c}{Self-training based open-vocabulary detection methods} \\
\noalign{\smallskip}
\hline
\noalign{\smallskip}
OVD-ALBEF\cite{gao2021towards}   & ALBEF & \cmark  & 30.8 & {\color{gray}46.1} & {\color{gray}42.1}\\
Detic\cite{zhou2022detecting} & - & \xmark & 27.8 & {\color{gray}47.1} & {\color{gray}45.0} \\
\rowcolor{mypink} VTP-OVD (w/o FT)   &CLIP & \xmark & 29.8 & {\color{gray}51.8}& {\color{gray}46.1}\\\rowcolor{mypink} 
VTP-OVD (Ours)  & CLIP & \xmark & \textbf{31.5}& {\color{gray}51.9}& {\color{gray}46.6}      \\
\noalign{\smallskip}
\hline
\end{tabular}
\end{sc}
\end{small}
\end{center}
\vspace{-5mm}
\end{table*}

\begin{table}
\caption{Generalization performance when adopted to other detection datasets. Note that all the models are only trained on the base classes of COCO dataset and mAP(\%) is reported under all classes.}
\vspace{-1mm}
\label{tab:generalization}
\begin{center}
\begin{small}
\begin{sc}
\setlength\tabcolsep{3pt} 
\begin{tabular}{lccccr}
\hline
\noalign{\smallskip}
Method & VOC & Objects365 & LVIS & COCO \\
\noalign{\smallskip}
\hline
\noalign{\smallskip}
OVR-CNN  & 52.9 & 4.6 & 5.2 & 39.9 \\
OVD-ALBEF  & 59.2& 6.9 & 8 & 42.1 \\
\rowcolor{mypink}
Ours     & \textbf{61.1} &  \textbf{7.4} & \textbf{10.6} & \textbf{46.6}  \\
\noalign{\smallskip}
\hline
\end{tabular}
\end{sc}
\end{small}
\end{center}
\end{table}

\begin{table}[t!]
\makeatletter\def\@captype{table}
\caption{Ablation Study for different prompt settings on the novel classes of COCO dataset. H.P, T.P, V.P, S.T denote the human pre-defined prompt, text prompt, vision prompt and self-training stage. The mAP(\%) of pseudo labels and the mIoU(\%) of dense score map on novel classes are reported.}
\setlength\tabcolsep{2.5pt} 
\begin{center}
\begin{small}
\begin{sc}
\begin{tabular}{cccc|cc}
\hline
\noalign{\smallskip}
  H.P & T.P & V.P & S.T & mAP(\%) & mIoU (\%) \\
\noalign{\smallskip}
\hline
\noalign{\smallskip}
\xmark &  \xmark  &  \xmark  & \xmark& 22.5& 30.9  \\
\cmark& \xmarkg   &  \xmarkg  &\xmarkg &23.7$^{+1.2}$ & 31.1$^{+0.2}$ \\
\xmarkg&\cmark   & \xmarkg &\xmarkg & 25.4$^{+2.9}$& 31.3$^{+0.4}$\\
\xmarkg&\xmarkg&\cmark  &\xmarkg & 25.1$^{+2.6}$& 33.1$^{+2.2}$\\
\xmarkg&\cmark & \cmark&\xmarkg & 26.0$^{+3.5}$& \textbf{35.4$^{+4.5}$} \\\rowcolor{mypink}
\xmarkg&\cmark & \cmark &\cmark & \textbf{31.5$^{+9.0}$}& /       \\
\noalign{\smallskip}
\hline
\end{tabular}
\end{sc}
\end{small}
\end{center}
\label{tab:prompt}
\end{table}

\subsection{Implementation Details}
\label{sec:implement}

We utilize a pre-trained CLIP~(RN50$\times $16)~\cite{radford2021learning} model
 as the VLM.
Note that our method is compatible with conventional VL models. For simplicity, we take CLIP as an example.
All the detection models are implemented on the mmdetection~\cite{mmdetection}
codebase and follow the default setting as Mask-RCNN~(ResNet50)~\cite{he2017mask}, 1X schedule unless otherwise mentioned.
At the adapting stage, the prompt-enhanced CLIP is trained for 5 epochs, with the text prompt learning rate set to 1e-1 and the visual prompt set to 1e-5 separately. 
For the pseudo labeling stage, the RPN is trained only on the base classes in the 2X schedule.
The objectness score threshold for RPN is set to 0.98.
The similarity threshold $\delta$ and score threshold $\gamma$ are set to 0.6 and 0.4.
For the self-training stage, 
following the default setting, 
we adopt a Mask-RCNN~(ResNet50)~\cite{he2017mask} with the last classification layer replaced by class embeddings output by the text encoder.
We train the detector for 12 epochs, with the learning rate decreased by a factor of 0.1 at 8 and 11 epochs with 8 V100 GPUs.
The initial learning rate is set to 0.04 with batch size 32, and the weight decay is set to 1e-4.
We extract the categories of large-scale object detection dataset~(i.e., LVIS~\cite{gupta2019lvis} and Objects365~\cite{shao2019objects365}) as the combined categories set $C$, which contains about 1k categories. We generate non-base pseudo labels for training, while evaluate only on novel categories and base categories following~\cite{zareian2021open}.

\begin{table}[t!]
\caption{Ablation study for different structure of text prompt. Embedding, LSTM and MLP denote the learnable embeddings, a bidirectional  LSTM and a 
two-layer multilayer perceptron. The mAP(\%) of pseudo labels are reported.}
\begin{center}
\begin{tabular}{ccc|cc}
\hline
\noalign{\smallskip}
  Embedding & LSTM & MLP & mAP(\%) & \\ 
\noalign{\smallskip}
\hline
\noalign{\smallskip}
 \xmark& \xmark   & \xmark    & 22.5  \\ 
\cmark&\cmark   & \xmarkg & 23.6$^{+1.1}$& \\
\cmark&\xmarkg& \cmark  & 24.3$^{+1.8}$& \\
\rowcolor{mypink}
\cmark & \cmark &\cmark& \textbf{25.4$^{+2.9}$}&       \\ 
\noalign{\smallskip}
\hline
\end{tabular}
\end{center}
\label{sample-table}
\end{table}


\vspace{-1mm}
\subsection{Main Results}
\label{sec:main_result}

\textbf{COCO Dataset.}
We compare our VTP-OVD with existing open-vocabulary methods on the COCO dataset~\cite{lin2014microsoft}. 
As shown in Table \ref{table:main_results}, VTP-OVD achieves the state-of-the-art performance~(i.e., 31.5\% mAP) on the novel categories of the COCO dataset and 4.5\% mAP improvement on overall categories compared to another self-training based method OVD-ALBEF~\cite{gao2021towards}.
Besides, by comparing with the VTP-OVD~(w/o FT), e.g., 31.5\% mAP vs. 29.8\% mAP, we further demonstrate the necessity of the fine-grained adapting stage with learnable visual and text prompts for dense alignment tasks.
Note that we utilize the basic vision-language model~(VLM) CLIP~\cite{radford2021learning} and do not use the caption data, guaranteeing the transfer ability to other VLMs or pre-training datasets without caption information.
Clarification should be made that self-training-based methods~(e.g., OVD-ALBEF~\cite{gao2021towards} and ours) usually achieve slightly worse on base classes compared to the knowledge-distillation~(KD) method~(e.g., ViLD~\cite{gu2021open}).
We attribute this to the reason that self-training methods make the model optimize more towards novel classes via generating massive pseudo labels on them, while KD-based methods try to keep the performance of the base class by only distilling the classification ability.

\textbf{Generalization Abilities.} To further demonstrate the open-vocabulary ability of the detector trained through the VTP-OVD pipeline, we directly transfer the final detection model trained on COCO datasets, to other detection datasets, including PASCAL VOC~\cite{everingham2010pascal}, Object365~\cite{shao2019objects365} and LVIS~\cite{gupta2019lvis}. 
Benefit from the training on non-base pseudo labels, experimental results in Table~\ref{tab:generalization} show that our VTP-OVD achieves the best generalization ability even adopted to the datasets with much more categories than the pre-trained dataset, i.e., 81 classes in COCO vs. 1203 classes in LVIS.
Note that we do not compare with ViLD~\cite{gu2021open} since it does not provide either this experiment result or the code.


\begin{figure}[h]
		\begin{center}

\includegraphics[ width=1.0\linewidth]{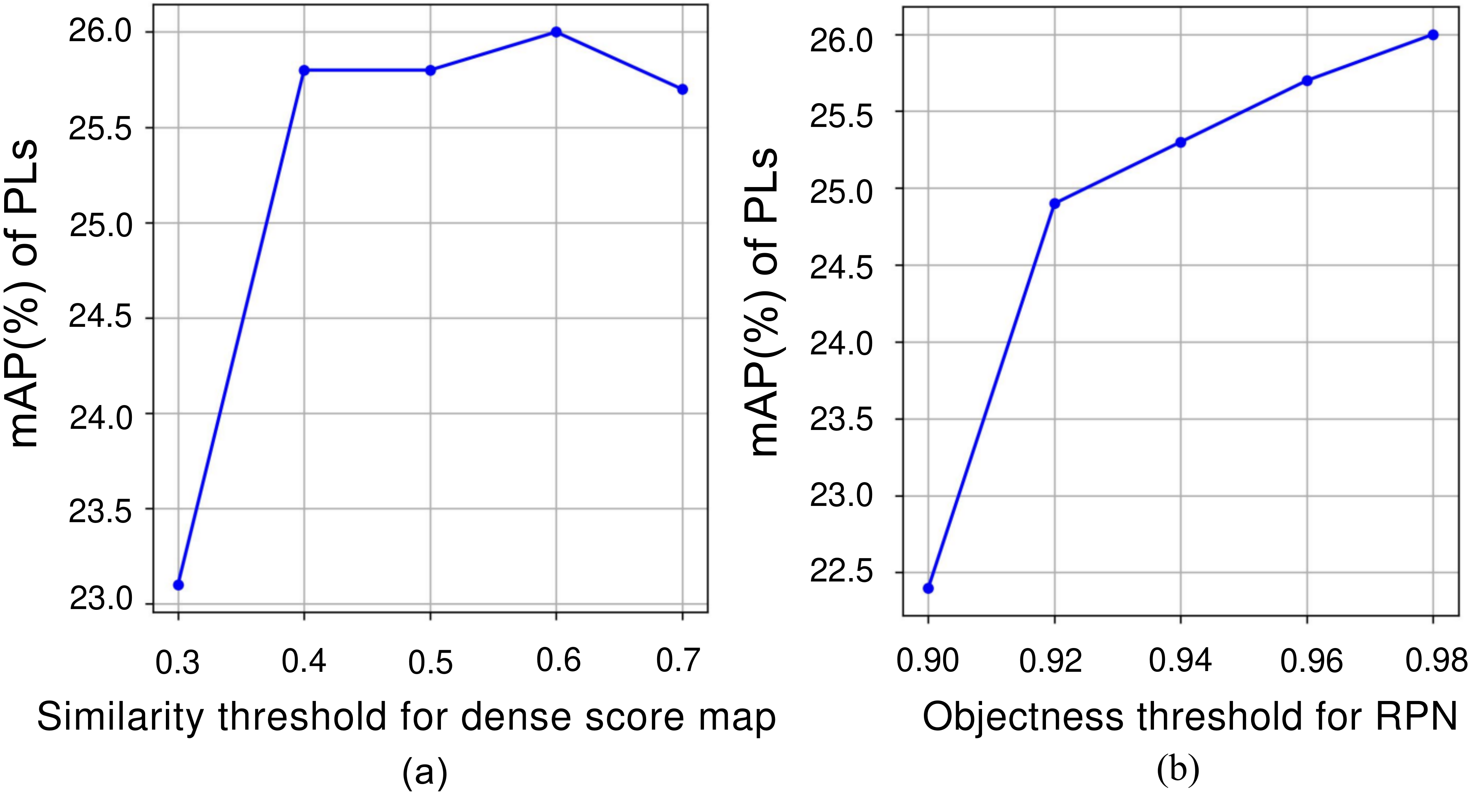}
		\end{center}
		\vspace{-4mm}
		\caption{Ablation studies of pseudo label generation on the (a) similarity threshold for dense score map (b) objectness threshold for RPN. `PLs' represents pseudo labels.}
		\label{fig:ablation_PL}
\end{figure}

\begin{figure*}[h!]
		\begin{center}
            \includegraphics[ width=1.0\linewidth]{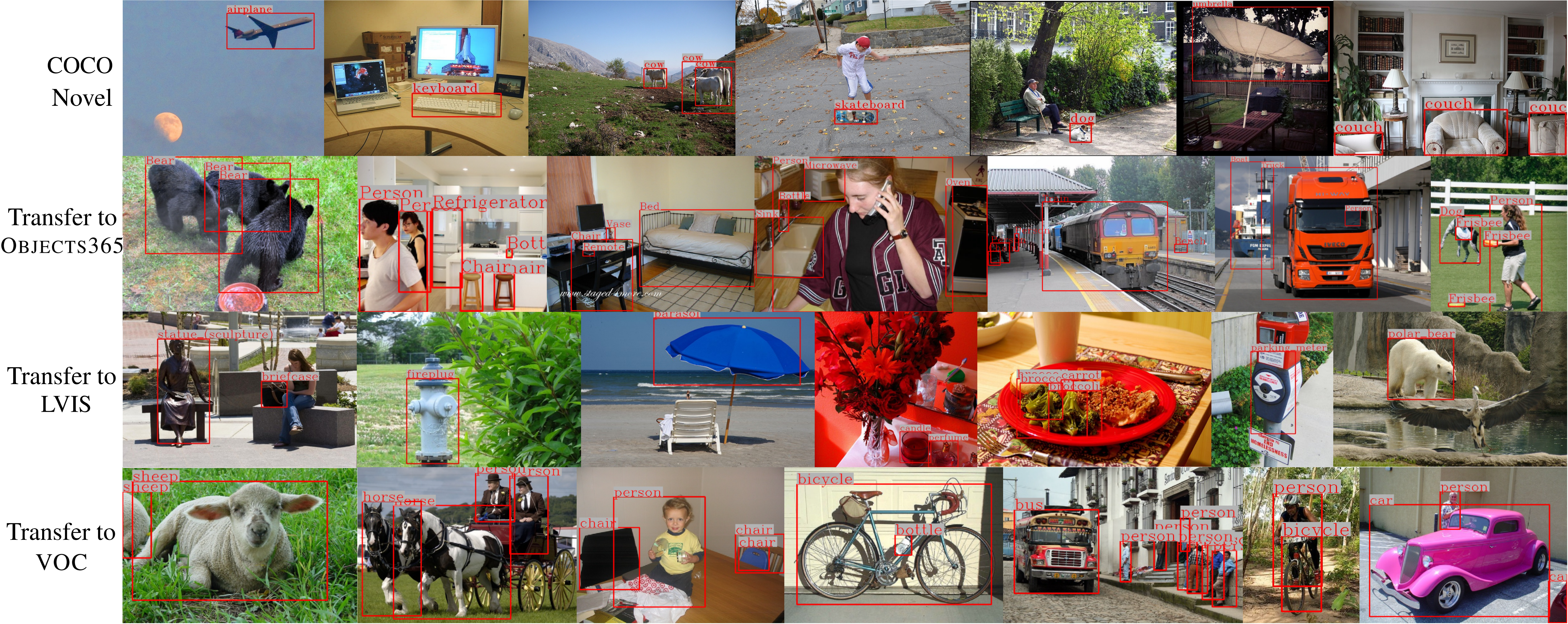}
		\end{center}
		\caption{Qualitative detection results on the novel classes of COCO dataset and the transfer performance on LVIS dataset. Note that the detection model is trained without any annotations of these categories.}
		\label{fig:novel detection result}
\end{figure*}

\begin{figure}[htbp]
		\begin{center}

\includegraphics[ width=1.0\linewidth]{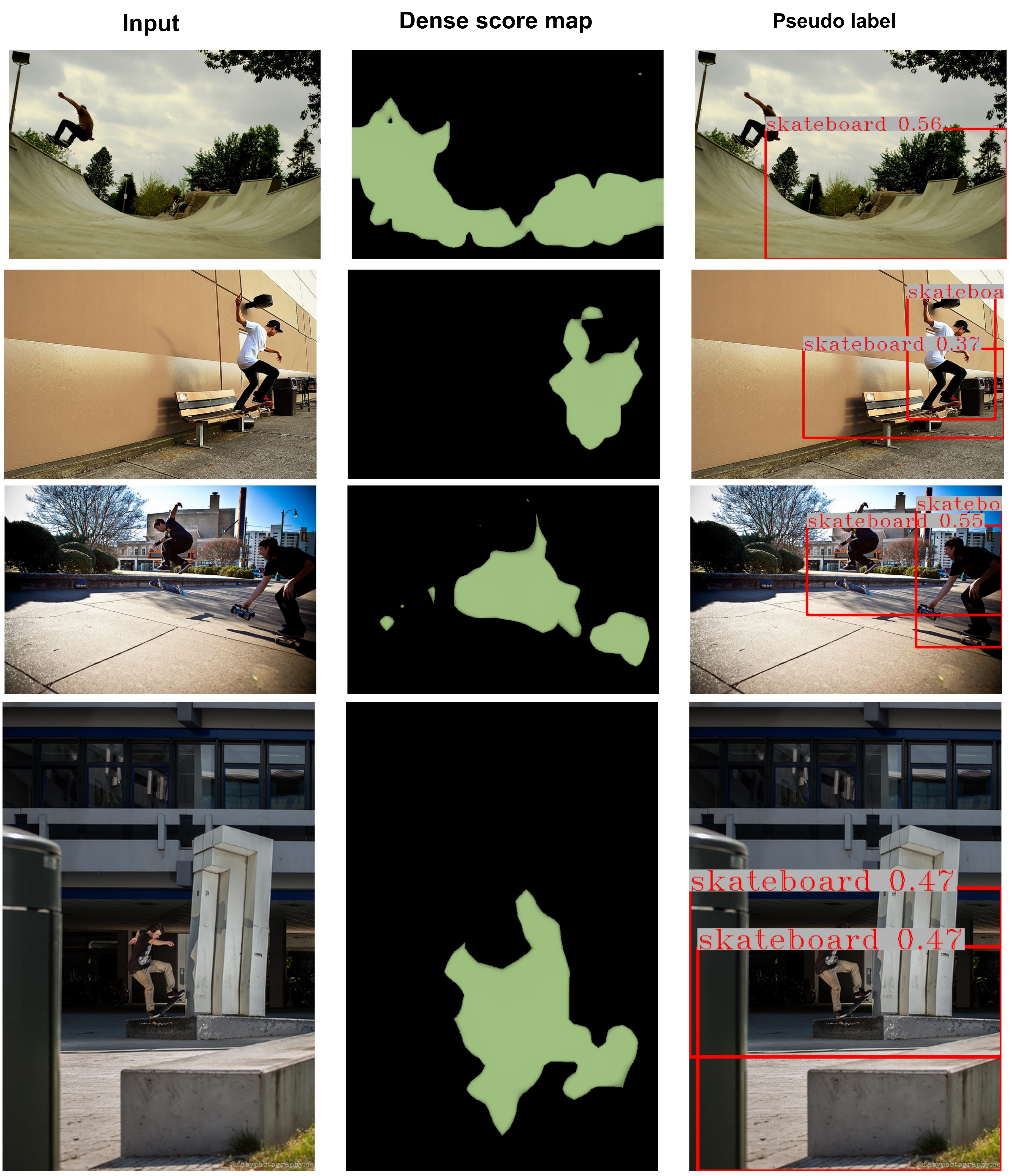}
		
		\caption{Illustration of the dense score map and the pseudo labels of class skateboard.}
		\vspace{-5mm}
		\label{fig:failure_cases}
		\end{center}
\end{figure}

\begin{figure*}[h!]
		\begin{center}
            \includegraphics[ width=0.95\linewidth]{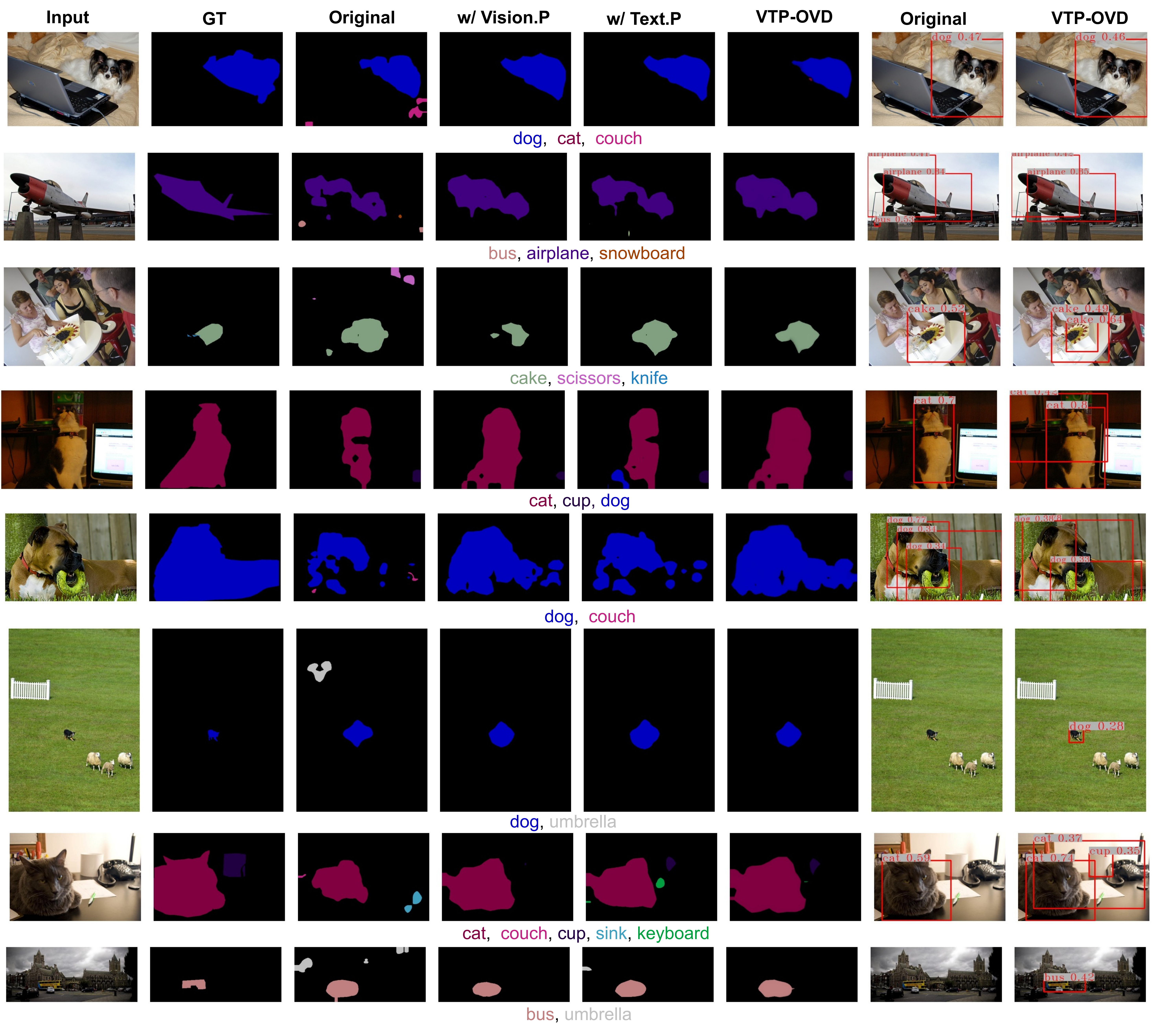}
		\end{center}
		\caption{Visualization of the learnable prompt modules' effect on dense alignment ability and the generated pseudo labels of novel classes on COCO dataset. The Vision.P and Text.P denote the vision prompt and text prompt modules, respectively.}
		\label{fig:segmentation result}
\end{figure*}

\subsection{Ablation Study}
\label{sec:ablations}

\paragraph{\textit{What's the effect of the learnable prompt modules for pseudo label generation?}}
We introduce two learnable prompt modules to 
obtain the dense alignment ability, which is important for pseudo bounding-box label generation.
To analyze the effect of proposed visual and text prompts, we conduct the quantitative experiments with different combinations of prompt modules to show the quality of pseudo labels~(with mAP) and dense score map~(with mIoU on 
novel classes).
As shown in Table \ref{tab:prompt}, adopting hand-crafted prompts improves the baseline~(without prompt) by 1.2\% mAP increment on pseudo labels' quality, while adding text and visual prompts separately gain additional 1.7\% and 1.4\% increment, respectively.
Combining visual and text prompts makes better performance~(+3.5\%) and the best dense score map~(35.4\% mIoU).
After adopting the 
self-training stage, it reaches the state-of-the-art detection performance~(31.5\% mAP) on COCO
novel classes.

\begin{table}[ht!]
\scriptsize
    \setlength{\tabcolsep}{10pt}
    \centering
    \caption{Ablation studies for category token position and prompt token number of text prompt.}
    \begin{tabular}{c|ccc}
    \toprule
     CATEGORY TOKEN POSITION & FRONT & MID & BACK\\
    \midrule
    ($L$,\ $M$) &(0,1) & (1,2) &  (1,1)\\
    mAP (\%) & 25.4 & 25.5  & \textbf{25.6}\\
    \midrule
    PROMPT TOKEN NUMBER & 1 & 3 & 5\\
    \midrule
    ($L$,\ $M$) &(1,1) & (3,3) & (5,5) \\
    mAP (\%) & \textbf{25.6}& 25.4 & 25.3\\
    \bottomrule
    \end{tabular}
    \label{ablation}
    \label{tab:lm_ablation}
\end{table}

\paragraph{\textit{What's the effect of different text prompt structures?}}
As shown in Table \ref{sample-table}, we evaluate the effect of different structures of only the text prompts.
Based on the input learned embeddings, separately using LSTM and MLP obtain a relatively small improvement due to the lack of the contextual information of prompt tokens. 
Adopting both LSTM and MLP for the association of prompt tokens achieves the best performance, i.e., 25.4\% mAP for generated pseudo labels on novel classes of COCO. 
The ablations of the category token position and prompt token number which are decided by $L$ and $M$ in Eq.4 are shown in Table~\ref{tab:lm_ablation}.




\paragraph{\textit{How the pseudo labeling robust to hyper parameters?}}
As shown in Fig.~\ref{fig:ablation_PL}, we conduct the ablation studies on the two most important hyper parameters, including the similarity threshold $\delta$ for dense score map to compute connected regions and objectness threshold for RPN to filter the low-confident proposals, in fine-tuning stage. 
Note that when we ablation one parameter, we keep the other fixed as the default value.
Observation can be made that our pseudo labeling strategy is robust to this two parameters and achieves the highest when similarity threshold and objectness threshold set to 0.6 and 0.98 respectively.
Low similarity and objectness threshold bring much false positives while high similarity and objectness threshold decrease the overall recall.





\subsection{Qualitative Results} 

\noindent\textbf{Qualitative examples.} We visualize the final detection performance of VTP-OVD on the novel classes of COCO dataset and the transfer performance on Objects365, LVIS, and Pascal VOC in Fig.~\ref{fig:novel detection result}.
We can observe that our VTP-OVD can obtain high-quality bounding-boxes for novel classes on COCO even without any annotation of these categories. Besides, as shown in the third row of Fig.~\ref{fig:novel detection result},  our VTP-OVD can detect rare classes like fireplug and polar bear, demonstrating its open-vocabulary detection ability.

However, we also find our VTP-OVD fails in detecting several novel categories, including skateboard, snowboard, etc, during the pseudo labeling stage and self-training stage. 
By observing the dense score map and the pseudo labels of skateboard class in Fig.~\ref{fig:failure_cases}, we attribute this phenomenon to the generation of wrong pseudo labels.

To explain the reason, we think that some objects are often related to the environment, such as skateboards, skateboarders, and skateboard playground often appear together, thus trained with the alignment based on global image information and text embeddings, current vision-language model usually cannot distinguish these objects from other objects in same scene.  

\noindent\textbf{Effect of prompt modules.} In Fig. \ref{fig:segmentation result}, we visualize the effect of proposed prompt modules on the generation of the pseudo labels of novel classes. 
By comparing the dense score maps in the third and sixth column
, observation can be made that through learnable visual-text prompts, the VLM can achieve a better dense alignment and improve the quality of pseudo labels.
Specifically, adopting the visual prompt tends to fill the dense alignment of the object, while adopting the text prompt usually explores the new classes or objects.
Combining these two prompts makes the better performance to generate more precise pseudo bounding-boxes. 

\begin{figure}[h!]
		\begin{center}
\includegraphics[ width=1.0\linewidth]{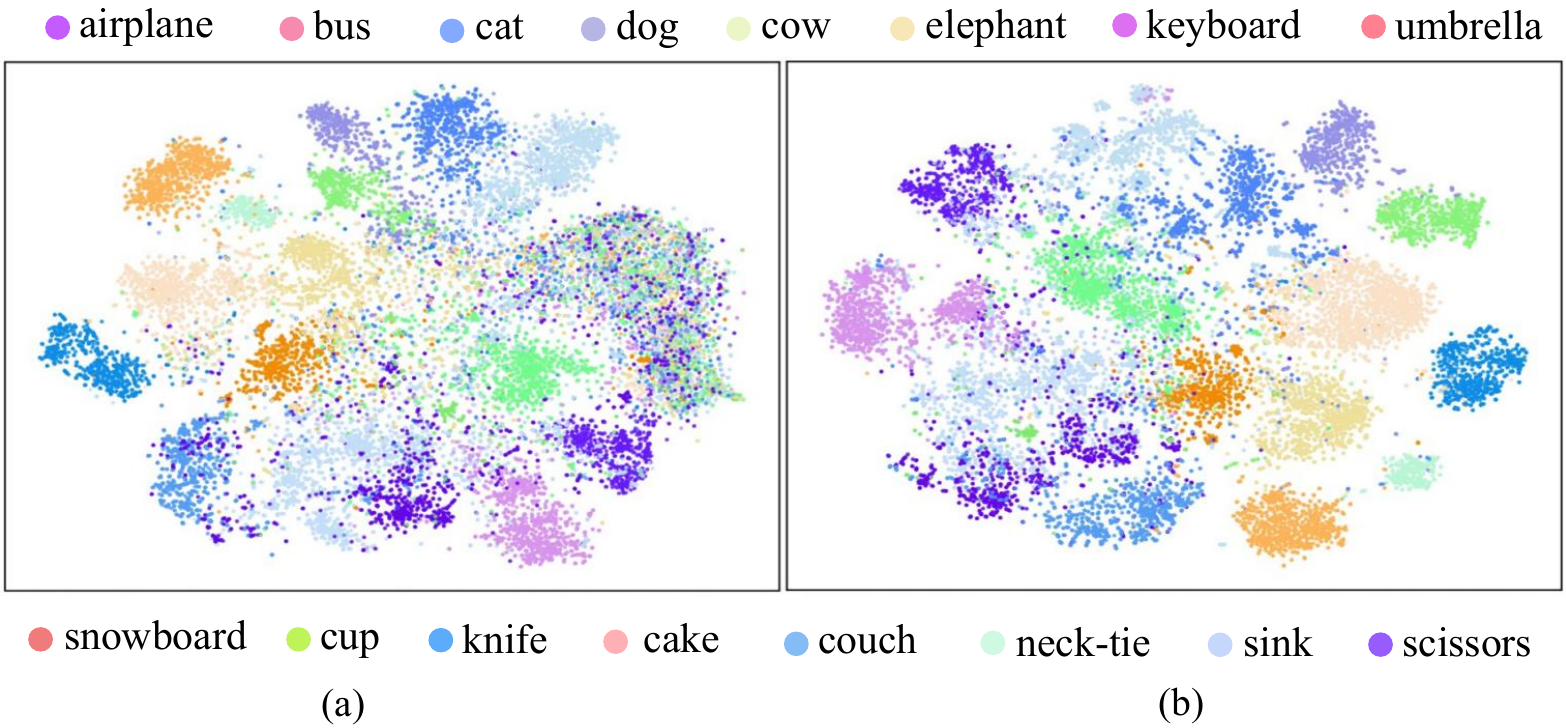}
		\end{center}
		\vspace{-4mm}
		\caption{Illustration of the visual embeddings generated by the (a) original CLIP and (b) CLIP with the proposed visual prompts via t-SNE~\cite{van2008visualizing}. Point in different color denotes the pixel belongs to different novel classes.}
		\vspace{-5mm}
		\label{fig:tsne}
\end{figure}

We also visualize the distribution of the pixel embeddings generated by the vision-language model CLIP (with or without learnable prompt modules) via t-SNE~\cite{van2008visualizing} on novel classes of COCO dataset in Fig.~\ref{fig:tsne}.
Observation can be made that adopting learnable prompt modules helps cluster the pixel embeddings in the same class and separate pixel embeddings from different categories, demonstrating their effects on dense alignments.



\section{Conclusions}

In this paper, we propose a novel open-vocabulary pipeline named VTP-OVD.
VTP-OVD introduces a new fine-tuning stage to enhance the self-training paradigm with dense alignments by adopting two learnable visual and text prompt modules.
Experimental results show that VTP-OVD achieves the state-of-the-art performance on the novel classes of COCO datasets and the best transfer performance when directly adapting the model trained on COCO to PASCAL VOC, Object365, and LVIS datasets.
Additional experiments also show that after fine-tuning with two learnable prompt modules, VTP-OVD obtains a more precise dense score map (4.3\% higher on mIoU) on novel classes. 
Nevertheless, VTP-OVD is a general pipeline of adopting the pre-trained vision-image encoder to dense prediction tasks, which can be easily extended to other tasks, e.g., open-vocabulary instance segmentation.
We hope VTP-OVD can serve as a strong baseline for future research on different open-vocabulary tasks.

\section{Acknowledments}
This work was supported in part by National Key R\&D Program of China under Grant No. 2020AAA0109700, Guangdong Outstanding Youth Fund (Grant No. 2021B1515020061), Shenzhen Science and Technology Program (Grant No. RCYX20200714114642083),
Shenzhen Fundamental Research Program(Grant No. JCYJ20190807154211365), Nansha Key RD Program under Grant No.2022ZD014 and Sun Yat-sen University under Grant No. 22lgqb38 and 76160-12220011. We thank MindSpore for the partial support of this work, which is a new deep learning computing framwork\footnote{https://www.mindspore.cn/}.

\ifCLASSOPTIONcaptionsoff
\newpage
\fi

\vspace{-0.2cm}
\bibliographystyle{IEEEtran}
\bibliography{main}	

\begin{thebibliography}{10}
\providecommand{\url}[1]{#1}
\csname url@samestyle\endcsname
\providecommand{\newblock}{\relax}
\providecommand{\bibinfo}[2]{#2}
\providecommand{\BIBentrySTDinterwordspacing}{\spaceskip=0pt\relax}
\providecommand{\BIBentryALTinterwordstretchfactor}{4}
\providecommand{\BIBentryALTinterwordspacing}{\spaceskip=\fontdimen2\font plus
\BIBentryALTinterwordstretchfactor\fontdimen3\font minus
  \fontdimen4\font\relax}
\providecommand{\BIBforeignlanguage}[2]{{%
\expandafter\ifx\csname l@#1\endcsname\relax
\typeout{** WARNING: IEEEtran.bst: No hyphenation pattern has been}%
\typeout{** loaded for the language `#1'. Using the pattern for}%
\typeout{** the default language instead.}%
\else
\language=\csname l@#1\endcsname
\fi
#2}}
\providecommand{\BIBdecl}{\relax}
\BIBdecl

\bibitem{everingham2010pascal}
M.~Everingham, L.~Van~Gool, C.~K. Williams, J.~Winn, and A.~Zisserman, ``The
  pascal visual object classes (voc) challenge,'' \emph{International journal
  of computer vision}, vol.~88, no.~2, pp. 303--338, 2010.

\bibitem{lin2014microsoft}
T.~Lin, M.~Maire, S.~J. Belongie, J.~Hays, P.~Perona, D.~Ramanan,
  P.~Doll{\'{a}}r, and C.~L. Zitnick, ``Microsoft {COCO:} common objects in
  context,'' in \emph{Computer Vision - {ECCV} 2014 - 13th European Conference,
  Zurich, Switzerland, September 6-12, 2014, Proceedings, Part {V}}, ser.
  Lecture Notes in Computer Science, D.~J. Fleet, T.~Pajdla, B.~Schiele, and
  T.~Tuytelaars, Eds., vol. 8693.\hskip 1em plus 0.5em minus 0.4em\relax
  Springer, 2014, pp. 740--755.

\bibitem{gupta2019lvis}
A.~Gupta, P.~Dollar, and R.~Girshick, ``Lvis: A dataset for large vocabulary
  instance segmentation,'' in \emph{Proceedings of the IEEE/CVF Conference on
  Computer Vision and Pattern Recognition}, 2019, pp. 5356--5364.

\bibitem{bansal2018zero}
A.~Bansal, K.~Sikka, G.~Sharma, R.~Chellappa, and A.~Divakaran, ``Zero-shot
  object detection,'' in \emph{Proceedings of the European Conference on
  Computer Vision (ECCV)}, 2018, pp. 384--400.

\bibitem{demirel2018zero}
B.~Demirel, R.~G. Cinbis, and N.~Ikizler-Cinbis, ``Zero-shot object detection
  by hybrid region embedding,'' \emph{arXiv preprint arXiv:1805.06157}, 2018.

\bibitem{rahman2019transductive}
S.~Rahman, S.~Khan, and N.~Barnes, ``Transductive learning for zero-shot object
  detection,'' in \emph{Proceedings of the IEEE/CVF International Conference on
  Computer Vision}, 2019, pp. 6082--6091.

\bibitem{hayat2020synthesizing}
N.~Hayat, M.~Hayat, S.~Rahman, S.~Khan, S.~W. Zamir, and F.~S. Khan,
  ``Synthesizing the unseen for zero-shot object detection,'' in
  \emph{Proceedings of the Asian Conference on Computer Vision}, 2020.

\bibitem{jia2021scaling}
C.~Jia, Y.~Yang, Y.~Xia, Y.-T. Chen, Z.~Parekh, H.~Pham, Q.~V. Le, Y.~Sung,
  Z.~Li, and T.~Duerig, ``Scaling up visual and vision-language representation
  learning with noisy text supervision,'' in \emph{International Conference on
  Machine Learning}, 2021.

\bibitem{chen2020uniter}
Y.~Chen, L.~Li, L.~Yu, A.~E. Kholy, F.~Ahmed, Z.~Gan, Y.~Cheng, and J.~Liu,
  ``{UNITER:} universal image-text representation learning,'' in \emph{Computer
  Vision - {ECCV} 2020 - 16th European Conference, Glasgow, UK, August 23-28,
  2020, Proceedings, Part {XXX}}, ser. Lecture Notes in Computer Science,
  A.~Vedaldi, H.~Bischof, T.~Brox, and J.~Frahm, Eds., vol. 12375.\hskip 1em
  plus 0.5em minus 0.4em\relax Springer, 2020, pp. 104--120.

\bibitem{radford2021learning}
A.~Radford, J.~W. Kim, C.~Hallacy, A.~Ramesh, G.~Goh, S.~Agarwal, G.~Sastry,
  A.~Askell, P.~Mishkin, J.~Clark \emph{et~al.}, ``Learning transferable visual
  models from natural language supervision,'' \emph{arXiv preprint
  arXiv:2103.00020}, 2021.

\bibitem{gao2021towards}
M.~Gao, C.~Xing, J.~C. Niebles, J.~Li, R.~Xu, W.~Liu, and C.~Xiong, ``Towards
  open vocabulary object detection without human-provided bounding boxes,''
  \emph{arXiv preprint arXiv:2111.09452}, 2021.

\bibitem{zhou2022detecting}
X.~Zhou, R.~Girdhar, A.~Joulin, P.~Kr{\"a}henb{\"u}hl, and I.~Misra,
  ``Detecting twenty-thousand classes using image-level supervision,''
  \emph{arXiv preprint arXiv:2201.02605}, 2022.

\bibitem{feng2022promptdet}
C.~Feng, Y.~Zhong, Z.~Jie, X.~Chu, H.~Ren, X.~Wei, W.~Xie, and L.~Ma,
  ``Promptdet: Expand your detector vocabulary with uncurated images,''
  \emph{arXiv preprint arXiv:2203.16513}, 2022.

\bibitem{liu2021pre}
P.~Liu, W.~Yuan, J.~Fu, Z.~Jiang, H.~Hayashi, and G.~Neubig, ``Pre-train,
  prompt, and predict: A systematic survey of prompting methods in natural
  language processing,'' \emph{arXiv preprint arXiv:2107.13586}, 2021.

\bibitem{zhou2021learning}
K.~Zhou, J.~Yang, C.~C. Loy, and Z.~Liu, ``Learning to prompt for
  vision-language models,'' \emph{arXiv preprint arXiv:2109.01134}, 2021.

\bibitem{shao2019objects365}
S.~Shao, Z.~Li, T.~Zhang, C.~Peng, G.~Yu, X.~Zhang, J.~Li, and J.~Sun,
  ``Objects365: A large-scale, high-quality dataset for object detection,'' in
  \emph{Proceedings of the IEEE/CVF International Conference on Computer
  Vision}, 2019, pp. 8430--8439.

\bibitem{9153181}
C.~Yan, Q.~Zheng, X.~Chang, M.~Luo, C.-H. Yeh, and A.~G. Hauptman,
  ``Semantics-preserving graph propagation for zero-shot object detection,''
  \emph{IEEE Transactions on Image Processing}, vol.~29, pp. 8163--8176, 2020.

\bibitem{yan2022semantics}
C.~Yan, X.~Chang, M.~Luo, H.~Liu, X.~Zhang, and Q.~Zheng, ``Semantics-guided
  contrastive network for zero-shot object detection,'' \emph{IEEE Transactions
  on Pattern Analysis and Machine Intelligence}, 2022.

\bibitem{9812473}
S.~Rahman, S.~Khan, and N.~Barnes, ``Polarity loss: Improving visual-semantic
  alignment for zero-shot detection,'' \emph{IEEE Transactions on Neural
  Networks and Learning Systems}, pp. 1--13, 2022.

\bibitem{li2021align}
J.~Li, R.~Selvaraju, A.~Gotmare, S.~Joty, C.~Xiong, and S.~C.~H. Hoi, ``Align
  before fuse: Vision and language representation learning with momentum
  distillation,'' \emph{Advances in Neural Information Processing Systems},
  vol.~34, 2021.

\bibitem{zareian2021open}
A.~Zareian, K.~D. Rosa, D.~H. Hu, and S.-F. Chang, ``Open-vocabulary object
  detection using captions,'' in \emph{Proceedings of the IEEE/CVF Conference
  on Computer Vision and Pattern Recognition}, 2021, pp. 14\,393--14\,402.

\bibitem{gu2021open}
X.~Gu, T.-Y. Lin, W.~Kuo, and Y.~Cui, ``Open-vocabulary object detection via
  vision and language knowledge distillation,'' \emph{arXiv preprint
  arXiv:2104.13921}, vol.~2, 2021.

\bibitem{bangalath2022bridging}
H.~Bangalath, M.~Maaz, M.~U. Khattak, S.~H. Khan, and F.~Shahbaz~Khan,
  ``Bridging the gap between object and image-level representations for
  open-vocabulary detection,'' \emph{Advances in Neural Information Processing
  Systems}, vol.~35, pp. 33\,781--33\,794, 2022.

\bibitem{selvaraju2017grad}
R.~R. Selvaraju, M.~Cogswell, A.~Das, R.~Vedantam, D.~Parikh, and D.~Batra,
  ``Grad-cam: Visual explanations from deep networks via gradient-based
  localization,'' in \emph{Proceedings of the IEEE international conference on
  computer vision}, 2017, pp. 618--626.

\bibitem{9709514}
Z.~Gu, S.~Zhou, L.~Niu, Z.~Zhao, and L.~Zhang, ``From pixel to patch:
  Synthesize context-aware features for zero-shot semantic segmentation,''
  \emph{IEEE Transactions on Neural Networks and Learning Systems}, pp. 1--15,
  2022.

\bibitem{li2019visualbert}
L.~H. Li, M.~Yatskar, D.~Yin, C.-J. Hsieh, and K.-W. Chang, ``Visualbert: A
  simple and performant baseline for vision and language,'' Preprint
  arXiv:1908.03557, 2019.

\bibitem{lin2021m6}
J.~Lin, R.~Men, A.~Yang, C.~Zhou, M.~Ding, Y.~Zhang, P.~Wang, A.~Wang,
  L.~Jiang, X.~Jia \emph{et~al.}, ``M6: A chinese multimodal pretrainer,''
  Preprint arXiv:2103.00823, 2021.

\bibitem{ramesh2021zeroshot}
A.~Ramesh, M.~Pavlov, G.~Goh, S.~Gray, C.~Voss, A.~Radford, M.~Chen, and
  I.~Sutskever, ``Zero-shot text-to-image generation,'' Preprint
  arXiv:2102.12092, 2021.

\bibitem{zhang2021ernie}
H.~Zhang, W.~Yin, Y.~Fang, L.~Li, B.~Duan, Z.~Wu, Y.~Sun, H.~Tian, H.~Wu, and
  H.~Wang, ``Ernie-vilg: Unified generative pre-training for bidirectional
  vision-language generation,'' \emph{arXiv preprint arXiv:2112.15283}, 2021.

\bibitem{8331856}
K.~Fu, J.~Li, J.~Jin, and C.~Zhang, ``Image-text surgery: Efficient concept
  learning in image captioning by generating pseudopairs,'' \emph{IEEE
  Transactions on Neural Networks and Learning Systems}, vol.~29, no.~12, pp.
  5910--5921, 2018.

\bibitem{8334194}
Z.~Yu, J.~Yu, C.~Xiang, J.~Fan, and D.~Tao, ``Beyond bilinear: Generalized
  multimodal factorized high-order pooling for visual question answering,''
  \emph{IEEE Transactions on Neural Networks and Learning Systems}, vol.~29,
  no.~12, pp. 5947--5959, 2018.

\bibitem{petroni2019language}
F.~Petroni, T.~Rockt{\"a}schel, P.~Lewis, A.~Bakhtin, Y.~Wu, A.~H. Miller, and
  S.~Riedel, ``Language models as knowledge bases?'' \emph{arXiv preprint
  arXiv:1909.01066}, 2019.

\bibitem{schick2020s}
T.~Schick and H.~Sch{\"u}tze, ``It's not just size that matters: Small language
  models are also few-shot learners,'' \emph{arXiv preprint arXiv:2009.07118},
  2020.

\bibitem{9881877}
R.~Mao, Q.~Liu, K.~He, W.~Li, and E.~Cambria, ``The biases of pre-trained
  language models: An empirical study on prompt-based sentiment analysis and
  emotion detection,'' \emph{IEEE Transactions on Affective Computing}, pp.
  1--11, 2022.

\bibitem{li2021prefix}
X.~L. Li and P.~Liang, ``Prefix-tuning: Optimizing continuous prompts for
  generation,'' \emph{arXiv preprint arXiv:2101.00190}, 2021.

\bibitem{yao2021cpt}
Y.~Yao, A.~Zhang, Z.~Zhang, Z.~Liu, T.-S. Chua, and M.~Sun, ``Cpt: Colorful
  prompt tuning for pre-trained vision-language models,'' 2021.

\bibitem{zhou2021denseclip}
C.~Zhou, C.~C. Loy, and B.~Dai, ``Denseclip: Extract free dense labels from
  clip,'' \emph{arXiv preprint arXiv:2112.01071}, 2021.

\bibitem{gao2021clip}
P.~Gao, S.~Geng, R.~Zhang, T.~Ma, R.~Fang, Y.~Zhang, H.~Li, and Y.~Qiao,
  ``Clip-adapter: Better vision-language models with feature adapters,''
  \emph{arXiv preprint arXiv:2110.04544}, 2021.

\bibitem{liu2021gpt}
X.~Liu, Y.~Zheng, Z.~Du, M.~Ding, Y.~Qian, Z.~Yang, and J.~Tang, ``Gpt
  understands, too,'' \emph{arXiv preprint arXiv:2103.10385}, 2021.

\bibitem{wang2021actionclip}
M.~Wang, J.~Xing, and Y.~Liu, ``Actionclip: A new paradigm for video action
  recognition,'' \emph{arXiv preprint arXiv:2109.08472}, 2021.

\bibitem{ren2015faster}
S.~Ren, K.~He, R.~Girshick, and J.~Sun, ``Faster r-cnn: Towards real-time
  object detection with region proposal networks,'' \emph{Advances in neural
  information processing systems}, vol.~28, pp. 91--99, 2015.

\bibitem{ghiasi2021simple}
G.~Ghiasi, Y.~Cui, A.~Srinivas, R.~Qian, T.-Y. Lin, E.~D. Cubuk, Q.~V. Le, and
  B.~Zoph, ``Simple copy-paste is a strong data augmentation method for
  instance segmentation,'' in \emph{Proceedings of the IEEE/CVF Conference on
  Computer Vision and Pattern Recognition}, 2021, pp. 2918--2928.

\bibitem{zhu2020don}
P.~Zhu, H.~Wang, and V.~Saligrama, ``Don't even look once: Synthesizing
  features for zero-shot detection,'' in \emph{Proceedings of the IEEE/CVF
  Conference on Computer Vision and Pattern Recognition}, 2020, pp.
  11\,693--11\,702.

\bibitem{rahman2020improved}
S.~Rahman, S.~Khan, and N.~Barnes, ``Improved visual-semantic alignment for
  zero-shot object detection,'' in \emph{Proceedings of the AAAI Conference on
  Artificial Intelligence}, vol.~34, no.~07, 2020, pp. 11\,932--11\,939.

\bibitem{mmdetection}
K.~Chen, J.~Wang, J.~Pang, Y.~Cao, Y.~Xiong, X.~Li, S.~Sun, W.~Feng, Z.~Liu,
  J.~Xu, Z.~Zhang, D.~Cheng, C.~Zhu, T.~Cheng, Q.~Zhao, B.~Li, X.~Lu, R.~Zhu,
  Y.~Wu, J.~Dai, J.~Wang, J.~Shi, W.~Ouyang, C.~C. Loy, and D.~Lin,
  ``{MMDetection}: Open mmlab detection toolbox and benchmark,'' \emph{arXiv
  preprint arXiv:1906.07155}, 2019.

\bibitem{he2017mask}
K.~He, G.~Gkioxari, P.~Doll{\'a}r, and R.~Girshick, ``Mask r-cnn,'' in
  \emph{Proceedings of the IEEE international conference on computer vision},
  2017, pp. 2961--2969.

\bibitem{van2008visualizing}
L.~Van~der Maaten and G.~Hinton, ``Visualizing data using t-sne.''
  \emph{Journal of machine learning research}, vol.~9, no.~11, 2008.

\end{thebibliography}
	
	%

\vspace{-5mm}
\begin{IEEEbiography}
[{\includegraphics[width=1in,height=1in,clip,keepaspectratio]{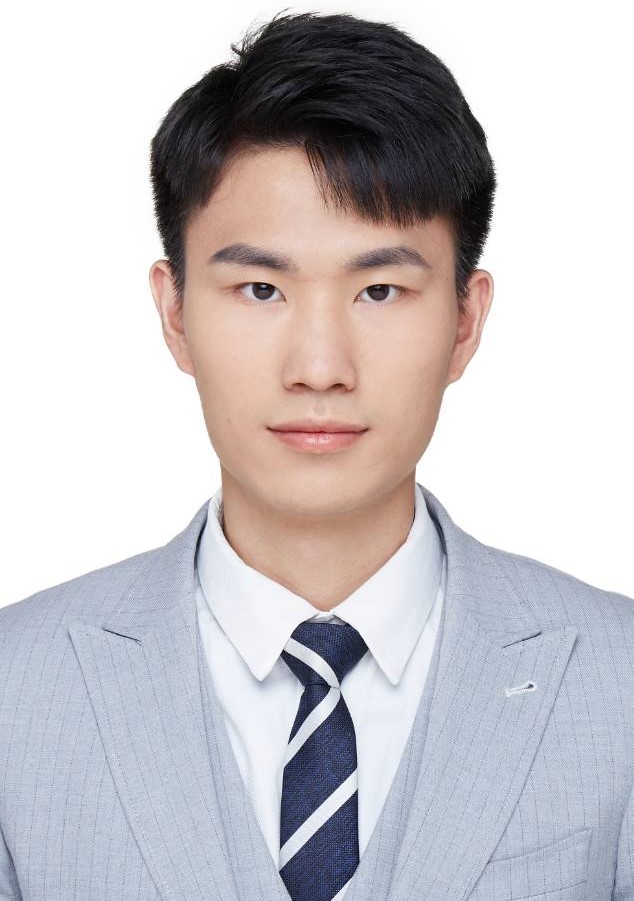}}]{Yanxin Long} is a third-year master in the School of Intelligent Systems Engineering, Sun Yat-Sen University. He works at the Human Cyber Physical Intelligence Integration Lab, Guangzhou under the supervision of Prof. Xiaodan Liang. Before that, He received his Bachelor Degree from the Comunication College, Xidian University in 2020. His research interests  include 2D detection and vision-and-language understanding.
\vspace{-15mm}
\end{IEEEbiography}

\begin{IEEEbiography}[{\includegraphics[width=1in,height=1in,clip,keepaspectratio]{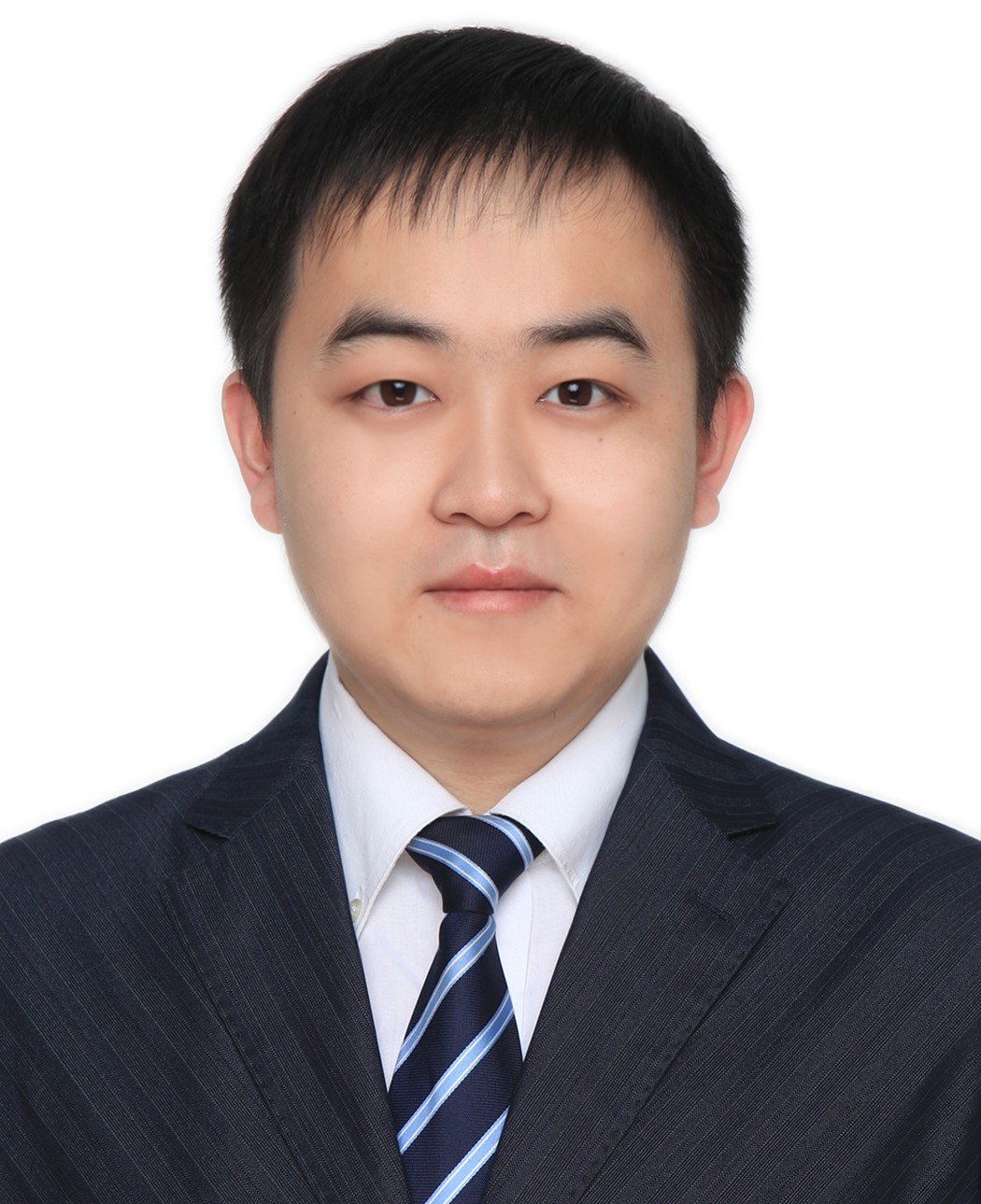}}]{Jianhua Han} received the Bachelor Degree 
 in 2016 and Master Degree in 2019 from Shanghai Jiao Tong University, China. He is currently a re-searcher with the Noahs Ark Laboratory, Huawei Technologies Co ., Ltd. His research interests lie primarily in deep learning and computer vision.
\vspace{-15mm}
\end{IEEEbiography}

\begin{IEEEbiography}[{\includegraphics[width=1in,height=1in,clip,keepaspectratio]{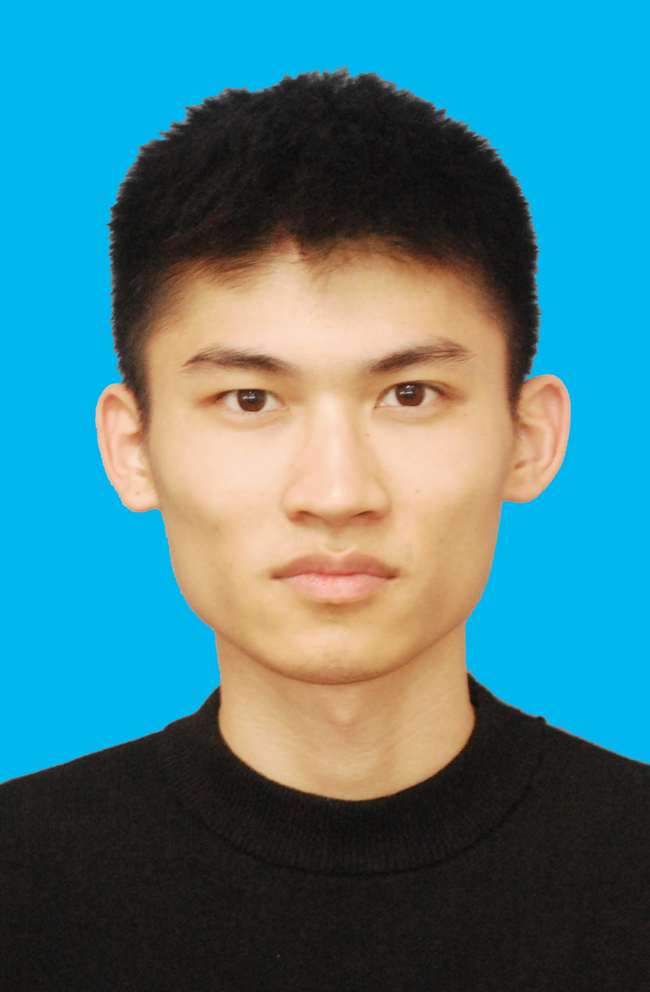}}]{Runhui Huang} is a two-year master in the School of Intelligent Systems Engineering, Sun Yat-sen University. He works at the Human Cyber Physical Intelligence Integration Lab, Guangzhou under the supervision of Prof. Xiaodan Liang. Before that, he received the Bachelor Degree from the School of Intelligent Systems Engineering, Sun Yat-sen University in 2021. His reasearch interests include vision-and-language pre-training.
\vspace{-15mm}
\end{IEEEbiography}

\begin{IEEEbiography}[{\includegraphics[width=1in,height=1in,clip,keepaspectratio]{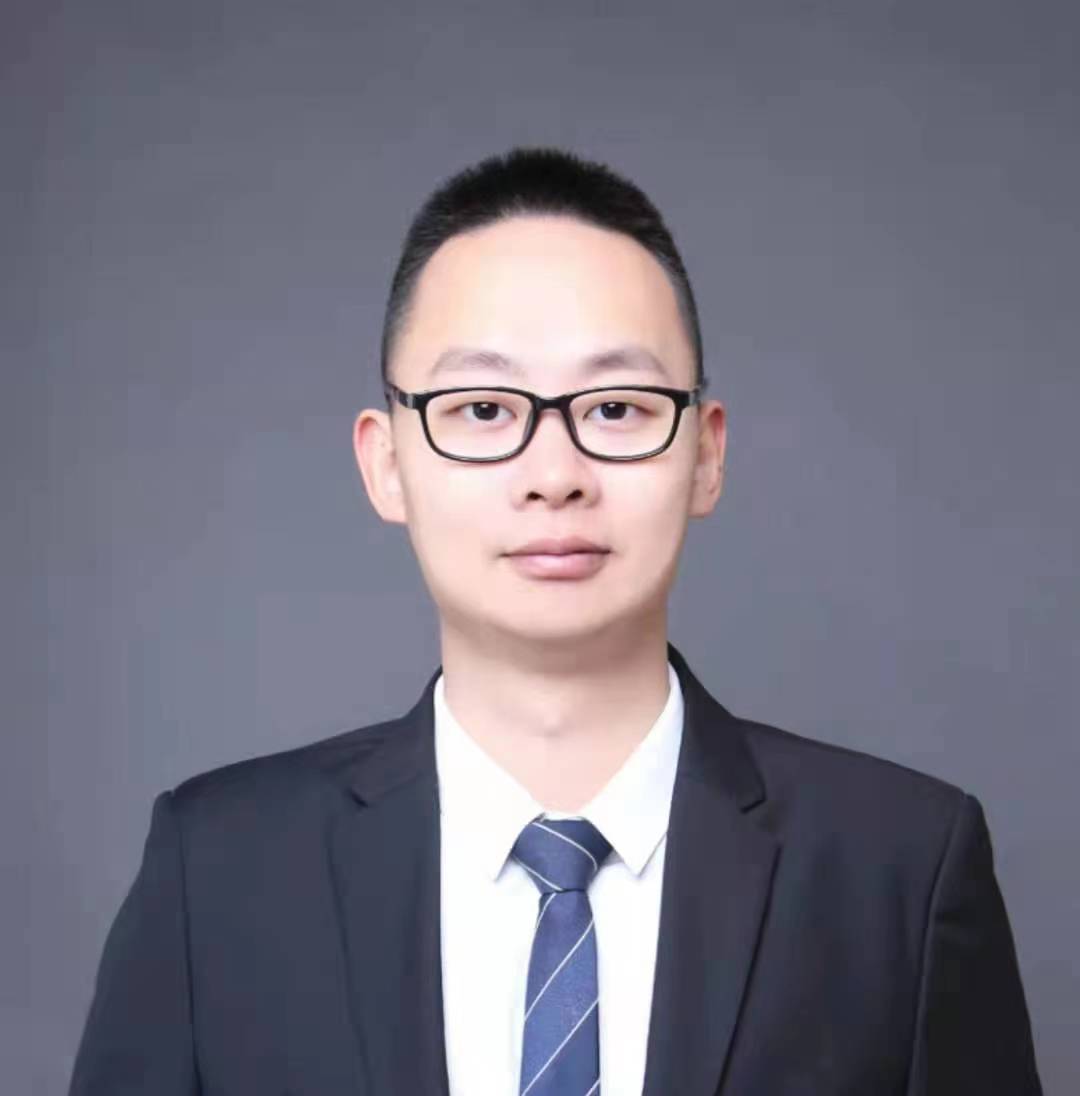}}]{Hang Xu} is currently a senior researcher in Huawei Noah Ark Lab. He received his BSc in Fudan University in 2012 and Ph.D. in Hong Kong University in Statistics in 2018. His research Interest includes machine Learning, vision-language models, object detection and AutoML. He has published 50+ papers in top-tier AI conferences such as NeurIPS, CVPR, ICCV, ICLR and some statistics journals, e.g. CSDA, Statistical Computing.
\vspace{-15mm}
\end{IEEEbiography}

\begin{IEEEbiography}[{\includegraphics[width=1in,height=1in,clip,keepaspectratio]{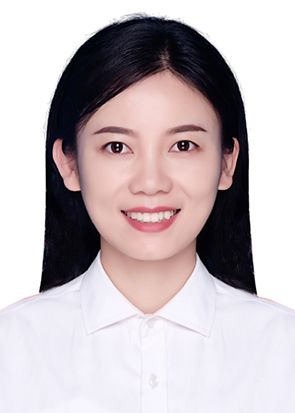}}]{Yi Zhu} received the B.S. degree in software engineering from Sun Yat-sen University, Guangzhou, China, in 2013. Since 2015, she has been a Ph.D. student in computer science at the School of Electronic, Electrical, and Communication Engineering, University of Chinese Academy of Sciences, Beijing, China. Her current research interests include object recognition, scene understanding, weakly supervised learning and visual reasoning.
\vspace{-15mm}
\end{IEEEbiography}

\begin{IEEEbiography}[{\includegraphics[width=1in,height=1in,clip,keepaspectratio]{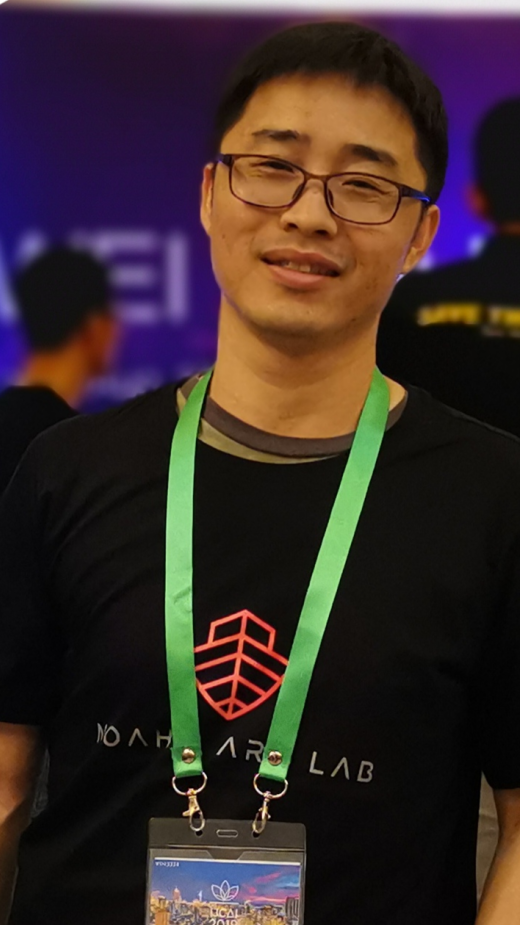}}]{Chunjing Xu} received his Bachelor's degree in Math from Wuhan University 1999, Master degree in Math from Peking University 2002, and Ph.D. from Chinese University of Hong Kong 2009. He was Assistant Professor and then Associate Professor at Shenzhen Institutes of Advanced Technology, Chinese Academy of Sciences. 
He became director of the computer vision lab in Noah's Ark lab, Central research institute in 2017. His main research interests focus on machine learning and computer vision.
\vspace{-15mm}
\end{IEEEbiography}

\begin{IEEEbiography}[{\includegraphics[width=1in,height=1in,clip,keepaspectratio]{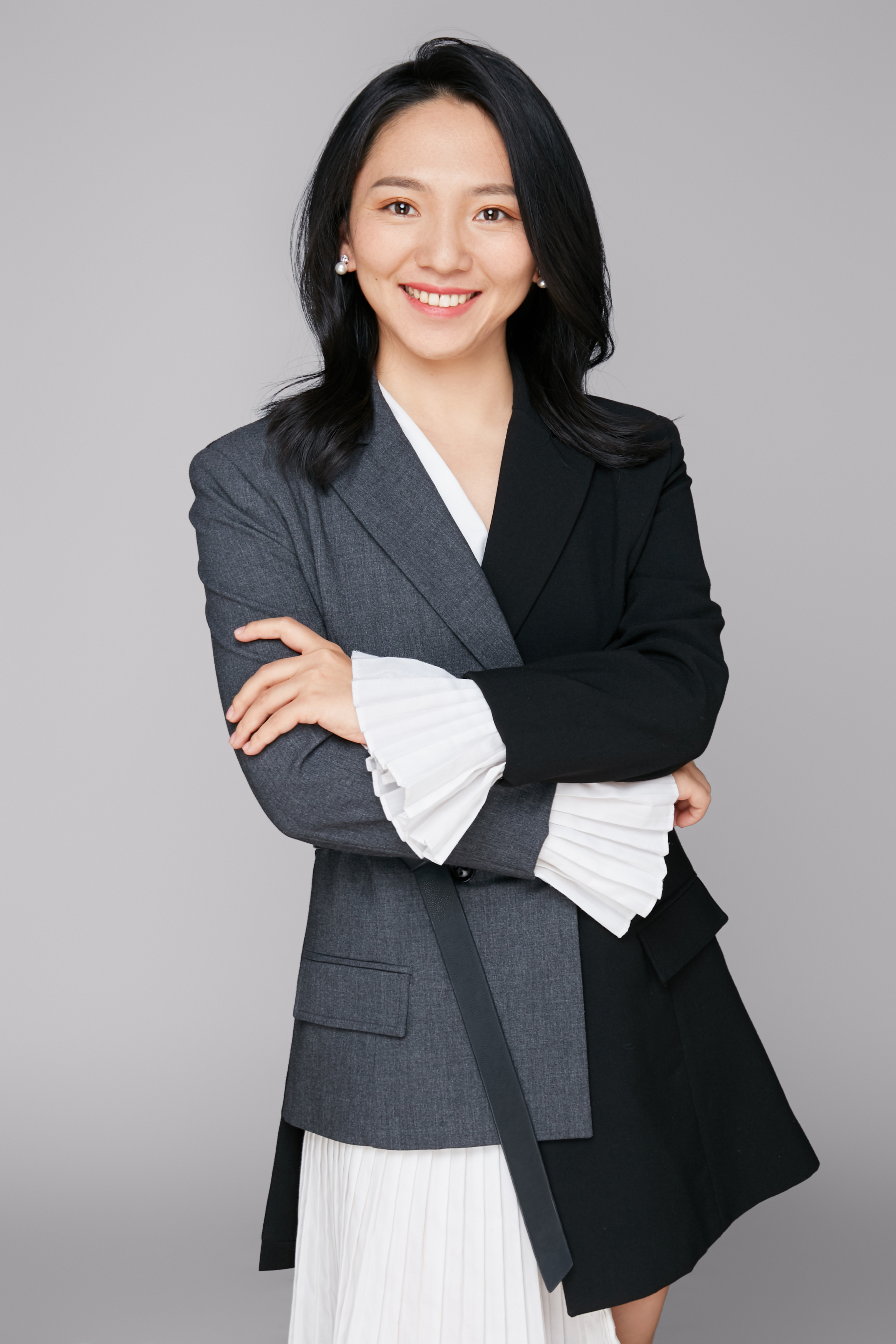}}]{Xiaodan Liang} is currently an Associate Professor at Sun Yat-sen University. She was a postdoc researcher in the machine learning department at Carnegie Mellon University, working with Prof. Eric Xing, from 2016 to 2018. She received her PhD degree from Sun Yat-sen University in 2016, advised by Liang Lin. She has published several cutting-edge projects on human-related analysis, including human parsing, pedestrian detection and instance segmentation, 2D/3D human pose estimation and activity recognition.
\vspace{-15mm}
\end{IEEEbiography}

\end{document}